\documentclass[lettersize,journal,twoside]{IEEEtran}

\usepackage{graphics}
\usepackage{epsfig}
\usepackage{mathptmx}
\usepackage{amsmath}
\usepackage{amssymb}
\usepackage{hyperref}
\usepackage[caption=false,font=normalsize,labelfont=sf,textfont=sf]{subfig}
\usepackage[table,xcdraw]{xcolor}
\usepackage{graphicx}
\usepackage{multirow}
\usepackage{pdflscape}
\usepackage{tablefootnote}
\usepackage{booktabs}
\usepackage[OT1]{fontenc} 

\usepackage{color}

\begin{document}

\title{\LARGE \bf
From Rigid to Soft Robotic Approaches for Minimally Invasive Neurosurgery
}

\author{Kieran Gilday$^1$, Irena Zubak$^2$, Andreas Raabe$^2$, Josie Hughes$^{1,*}$
\thanks{$^1$Authors with the CREATE-Lab, Institute of Mechanical Engineering, EPFL, Switzerland}
\thanks{$^{2}$Authors with the University Department of Neurosurgery, Inselspital Bern, Switzerland
        }%
\thanks{*Corresponding Author:  {\tt\small josie.hughes@epfl.ch}.}
\thanks{Manuscript submitted March 6, 2024}}

\markboth{Pre-print}%
{K. Gilday \MakeLowercase{\textit{et al.}}: From Rigid to Soft Robotic Approaches for Minimally Invasive Neurosurgery}

\maketitle

\begin{abstract}
Robotic assistance has significantly improved the outcomes of open microsurgery and rigid endoscopic surgery, however is yet to make an impact in flexible endoscopic neurosurgery. Some of the most common intracranial procedures for treatment of hydrocephalus and tumors stand to benefit from increased dexterity and reduced invasiveness offered by robotic systems that can navigate in the deep ventricular system of the brain. We review a spectrum of flexible robotic devices, from the traditional highly actuated approach, to more novel and bio-inspired mechanisms for safe navigation. For each technology, we identify the operating principle and are able to evaluate the potential for minimally invasive surgical applications. Overall, rigid-type continuum robots have seen the most development, however, approaches combining rigid and soft robotic principles into innovative devices, are ideally situated to address safety and complexity limitations after future design evolution. We also observe a number of related challenges in the field, from surgeon-robot interfaces to robot evaluation procedures. Fundamentally, the challenges revolve around a guarantee of safety in robotic devices with the prerequisites to assist and improve upon surgical tasks. With innovative designs, materials and evaluation techniques emerging, we see potential impacts in the next 5--10~years.
\end{abstract}

\begin{IEEEkeywords}
Robotic neurosurgery review, flexible endoscopy, soft surgical robotics.
\end{IEEEkeywords}

\section{Introduction}

\begin{figure}[t]
    \centering
    \includegraphics[width=0.99\columnwidth]{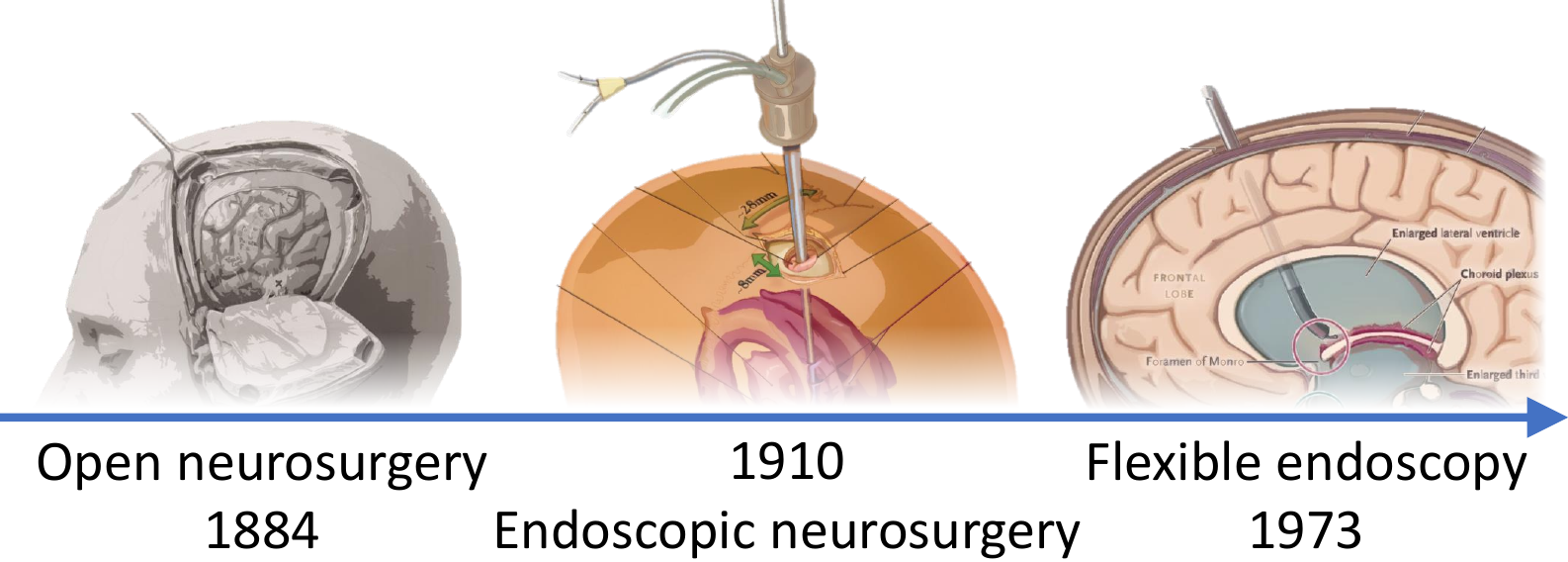}
    \caption{Types of neurosurgeries practiced today and their origin.}
    \label{fig:1}
\end{figure}

\IEEEPARstart{D}{espite} significant advancements in the field of medical robotics and improved understanding of brain anatomy, the majority of intercranial procedures are still performed via open surgery or rigid endoscopic surgery~\cite{elsabeh2021cranial,price2023using,zhao2015surgical,li2005neuroendoscopy}, Fig.~\ref{fig:1}. Neurosurgery has a long history, the first open surgery (where the skull is opened and brain issue is removed to access deeper sites) for brain tumor removal was performed in 1884. The first neurosurgical endoscopic procedure offering a less invasive approach was performed in 1910, and the first use of a flexible endoscope for ventriculostomies was reported in 1973~\cite{li2005neuroendoscopy,demerdash2017endoscopic}. Each of these techniques have seen significant refinement to modern neurosurgery, in particular the use of advanced imaging techniques and robotic assistance with microsurgery has significantly improved precision with enhanced visual feedback and mechanical advantage~\cite{nathoo2005touch,mcbeth2004robotics,faria2015review}. However, despite their promises of reduced risk, robotic flexible endoscopes have yet to see impact in neurosurgical applications outside of research environments~\cite{price2023using,zhong2020recent,gifari2019review}. In this article, we review a range of robot approaches and technologies that contribute towards creating more impactful, capable and safer neurosurgical tools for operating deep within the brain~\cite{rhoton2002lateral}.

Robotic assisted surgery can greatly increase the precision and dexterity at micro scales and reduce risks of errors~\cite{mcbeth2004robotics,zhao2015surgical}. This is particularly true with traditional robotics approaches, which emphasise rigid, usually metallic, materials with highly predictable motions~\cite{faria2015review}. Soft robotics is a more recent paradigm, which relaxes the strict control of rigid robotics and emphasises compliant materials such that their motions arise through environmental interactions~\cite{angrisani2019use,banerjee2018soft}. Whilst this approach and theory is somewhat exploited by existing flexible endoscopes~\cite{seah2018flexible,kume2016flexible}, this approach is not widely utilized, as evidenced by current commercial surgical device space~\cite{faria2015review,banerjee2018soft}. Whereas other reviews address limitations and challenges with specific approaches, e.g., rigid microsurgery~\cite{nathoo2005touch,doulgeris2015robotics}, soft devices for surgical applications~\cite{gifari2019review,zhong2020recent}, follow-the-leader (FTL) mechanisms~\cite{culmone2021follow}, flexible needle robots~\cite{siepel2021needle} and more~\cite{zhao2015surgical}, this review aims to more directly compare the different robotics approaches in ventriculostomy applications. By focusing on ventriculostomy we can more clearly define the robot criteria, and by directly comparing rigid and soft robotic approaches, as well as hybrid approaches combining ideas from both, we can better understand the challenges holding flexible robotic endoscopy back.

In order to address the question of robot technological suitability in minimally invasive neurosurgery we need to understand the task requirements. Endoscopic third ventriculostomy (ETV), biopsy and tumor resection are some of the most common procedures performed when operating within the ventricular system~\cite{kahle2016hydrocephalus,anderson2003surgical}. Flexible surgical devices allow for reduced invasiveness with fewer entry points, access to deep locations from entry points with the least risk, and the ability for surgeons to approach targets from multiple angles and inspect surrounding tissue~\cite{abbassy2018outcome}. In this review, we investigate how different types of robots try to address the requirements in these surgical tasks and what are the remaining challenges to move toward deployment and regulation of these technologies.  Observing the factors limiting the applications of these devices also allows us to see the wider challenges in the creation of minimally invasive flexible surgical robots that could also be applied to other challenging surgical applications.

Following an in-depth review of diverse surgical devices across rigid, soft and hybrid approaches, we select key representative technologies ranging from: segmented tendon driven designs with more than 10 degrees of freedom~\cite{gao2019continuum,henselmans2017memo} to the 3 degree of freedom flexible endoscope~\cite{ikeda2011evaluation}; those fabricated from metals~\cite{peyron2022magnetic} and much softer plastics~\cite{hawkes2017soft}; FTL motions through control~\cite{striegel2011determining,gao2019continuum} or automatic mechanisms~\cite{culmone2022memobox,kang2016first}, and more. We see significant limitations in pure soft robotics approaches, and despite their own limitations the most capable current devices are seen in the rigid robotics category. While currently exhibiting limitations in scale and flexibility, the hybrid approach offers highly innovative mechanisms with fewer moving parts and a large design space for future development. Overall, we identify challenges in guaranteeing safety in rigid robotic devices in the short term. In the longer term, we identify challenges in advancement of design and manufacturing techniques for novel hybrid robotics in order to solve small-scale, arbitrary path following while minimising the risk of damage from collisions along the entire length of the robot. Outside of technical challenges, we see the need for improvements in haptic feedback and surgical phantoms (physical or virtual replica of the tissue being studied/operated on) not only for surgeon training, but also robot design feedback and safety evaluation over diverse conditions.

In summary, this review defines the problem of robot assisted, minimally invasive neurosurgery, allowing us to give contributions in three areas. First, the surgical specifications to solve this problem from a robotics perspective. Secondly, a contrasting view of rigid, soft and hybrid robotic approaches at fulfilling these requirements. Finally, the remaining challenges in defining general and bespoke surgical specifications, and the technological challenges for the next generation of flexible neurosurgical devices.

In the remainder of this review, we first define the application and requirements for minimally invasive neurosurgery. In Section III, we introduce and compare the existing robot technologies for how well they meet each requirement. Section IV contains discussion on the remaining challenges, both in overcoming robot limitations and other factors for consideration in practical use. Finally, Section~V presents the conclusions.

\section{Requirements for Minimally Invasive Neurosurgery}

\noindent Neurosurgery requires interacting with delicate tissue where damage risks significant long-term health problems. This is especially the case deep within the brain~\cite{anderson2003surgical,rhoton2002lateral}. Safe access to these areas depends on more than just the control of a robot and avoiding sensitive areas~\cite{price2023using}. Other considerations must be made to minimise invasiveness, including: abrasion, risk of contamination, sources of damage other than excess contact pressure and the usability of the device.

\begin{figure}[t]
    \centering
    \includegraphics[width=0.99\columnwidth]{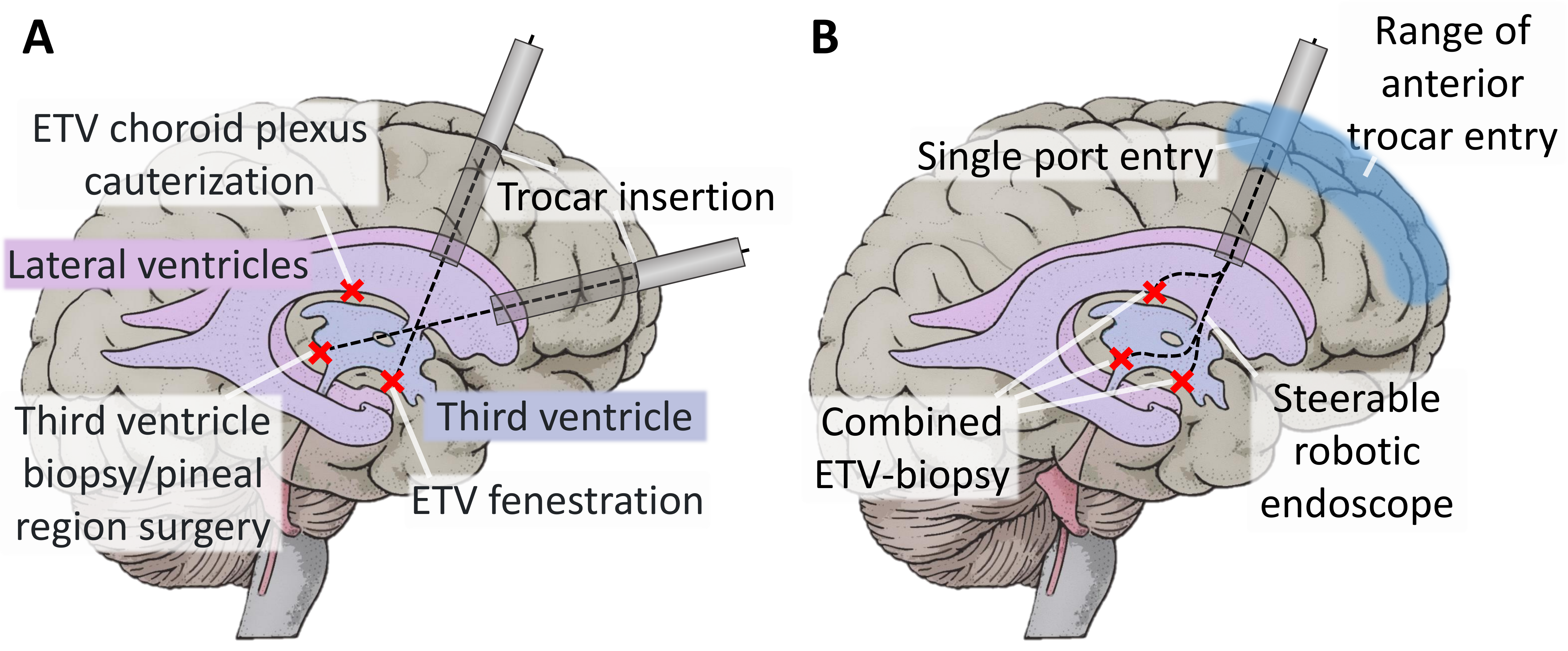}
    \caption{The ventricle system and surgical targets for different interventions. (A) Multi-port entry to access lateral ventricles and third ventricle. (B) Single-port entry with a flexible endoscope.}
    \label{fig:2}
\end{figure}

Example procedures that can benefit from robotic endoscopes are endoscopic third ventriculostomy (ETV) and intraventricular tumor surgery. These are two common neurosurgical procedures operating in the delicate and critical ventricle system deep within the brain~\cite{rhoton2002lateral}. Both of these are typically performed with rigid endoscopy, using multiple entry points to access different targets, and stand to benefit from robotic systems with increased flexibility~\cite{abbassy2018outcome}. The lateral ventricles and third ventricle, Fig.~\ref{fig:2}, are cavities deep within the brain filled with cerebrospinal fluid and can be navigated to access surgical targets.  In the following section we detail the requirements for common procedures within each ventricle, and also the general specifications for such robots. 

\subsection{Lateral ventricle interventions}

\noindent Procedures in the lateral ventricles include tumor surgery and cauterization of the choroid plexus (the tissue which overproduces cerebrospinal fluid in hydrocephalus cases). Additionally, the lateral ventricles are used as transventricular access to deep seated brain regions. The lateral ventricle walls contain many delicate structures and it is not well understood where and how much pressure can be applied before causing trauma and long-term negative health outcomes. For this reason, contact should be avoided wherever possible, except at the target site~\cite{rhoton2002lateral,kahle2016hydrocephalus}. Surgical targets are localised with preoperative medical imaging, allowing for pre-planning of trocar --- surgical tools for making puncture-like incisions/access channels, Fig.~\ref{fig:2} --- insertion for direct access with a rigid endoscope~\cite{abbassy2018outcome}. However, for surgeries with multiple target sites, multiple entry points are required, increasing risk. Even for surgeries with a single target, flexibility to access sites from entry points with less risk, from multiple angles or for inspecting surrounding tissues and blood vessels can improve outcomes~\cite{anderson2003surgical}.

Due to the small width of the lateral ventricles and long length, a flexible robot is required to have small bending radii ($<10$~mm) to access or inspect multiple locations approximately 30~mm in either direction from the access point. Significant total curvature is required, likely two $90^{\circ}$ turns. Few degrees of freedom are needed as even relatively simple paths can greatly increase the workspace of the robot within the ventricles.

\subsection{Third ventricle interventions}

\noindent Access to the third ventricle is required for tumor surgery in the pineal region, safe surgery of intraventricular craniopharyngiomas, tumors located in the thalamus, as well as surgery for obstructive hydrocephalus such as third ventriculostomy~\cite{kahle2016hydrocephalus,liu2021simultaneous}. This cavity can be reached directly or through the foramen of Monro (the thin channels connecting the lateral and third ventricles) when enlarged due to the increased pressure caused by hydrocephalus. The third ventricle has a more complex and smaller shape than the lateral ventricles, requiring more degrees of freedom to avoid obstacles (particularly when accessing through the foramen of Monro), precise follow-the-leader-motions to avoid contact during propagation and retraction, and manipulator diameters less than 3~mm~\cite{rhoton2002lateral}.

\subsection{Specifications for general ventricular surgeries}

\noindent Bespoke manipulators can be developed for each surgery, however there are significant benefits that arise from a more general-purpose surgical instrument. Having a single manipulator requires less training and reduces chances of human error, in addition to allowing combined surgeries when needed, e.g. simultaneous ETV and tumor biopsy, minimising the invasiveness of individual surgeries and multiple entry points. Taking the above case studies, we can now define a specification for an ideal flexible endoscopic robot operating within the ventricular system.

\begin{figure*}[t]
    \centering
    \includegraphics[width=0.99\textwidth]{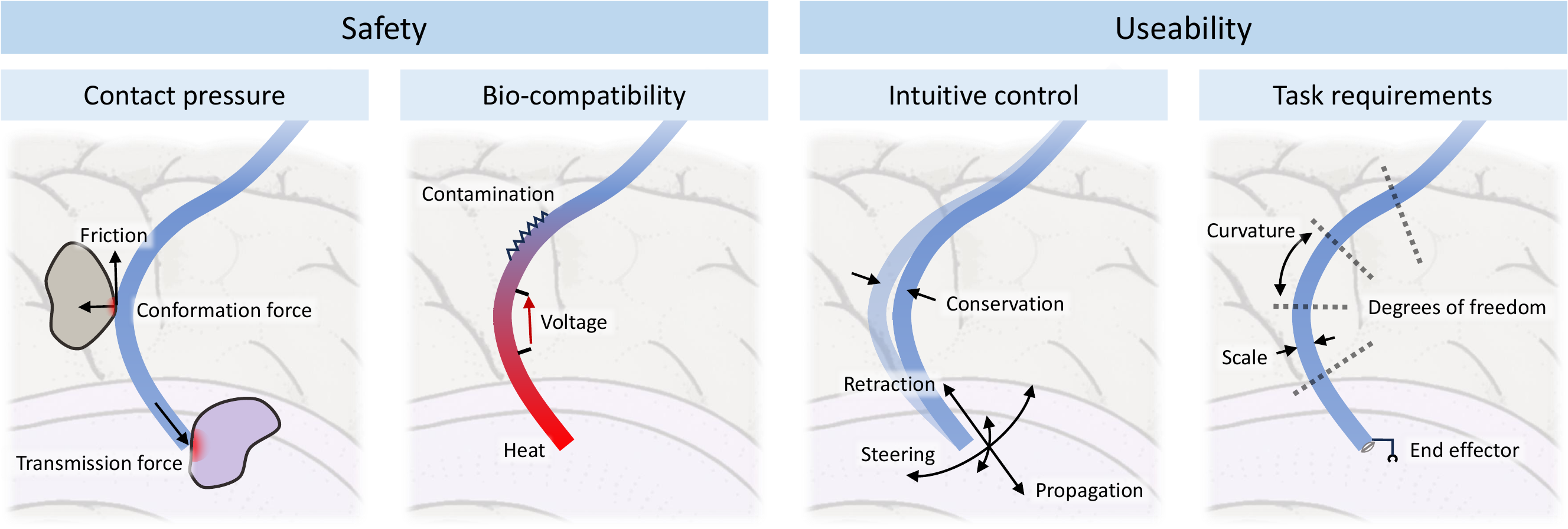}
    \caption{Requirements for effective minimally invasive neurosurgery. Safety: physical contact pressure and friction should be avoided except at the surgical target; the device should be bio-compatible and minimise risk of infection. Usability: follow-the-leader should be automatic, control should be intuitive for the surgeon; the device must meet the task requirements in terms of curvature, degrees of freedom, scale and end-effector capacity.}
    \label{fig:3}
\end{figure*}

Fig.~\ref{fig:3} illustrates the considerations for minimally invasive robotic neurosurgery. From the medical perspective, we must consider the safety of the device and any potential risks of different designs in order to improve patient outcomes. \emph{Physical contact} is key, which as discussed above, needs to be minimised, especially in critical regions where even light pressure can cause long-term health effects. This includes friction, where the insertion of the manipulator can cause abrasion or even cutting when steering around obstacles. Other risks include those from infection and other \emph{bio-incompatibilities}, which are more likely in complex manipulators which are difficult to sterilise and undesireable as single-use tools. Some manipulators are actuated internally with high voltages. Without suitable fail-safes, this has the potential to cause harm. This is similarly true for manipulators which are actuated by heat or generate significant heat energy.

From a technical perspective, the device usability and suitability for the neurosurgical use-cases are the main considerations. This includes the \emph{interface} between the device and the user (surgeon), where \emph{intuitive control} reduces risk of user error, automated follow-the-leader motions (steering happens only at the robot tip and as the robot propagates forwards the shape of the remainder of the robot should always conserve the path taken up to that point) minimise risk of contact, and direct/manual control can reduce risk in cases of robot failure. Finally, the actual \emph{surgical task specifications} need to be fulfilled, including the required curvature, number of degrees of freedom/turns required, the scale/width of the manipulator, and the carrying capacity of the end-effectors, internal optic fibres and other connections.

In summary:
\begin{itemize}
    \item \emph{Physical contact pressure} --- stress in brain tissue should be avoided, except at the surgical target. From estimates for brain concussion force, $\approx55$~kPa is enough to cause damage (equivalent to 0.5~N with a 3~mm tool). Delicate tissues on the ventricle walls may be damaged by even lower pressure. 
    \item \emph{Bio-compatibility} --- devices should meet regulatory requirements for sterility and withstand standard sterilization procedures. Internal voltages should be limited so no currents above 10~mA can be generated in brain tissue. Excess heat generation requires active cooling.
    \item \emph{Intuitive user control} --- FTL motions during propagation and retraction should be automatic with path deviation under 1~mm when close to a wall. Navigation interface should not exceed 4 degrees of control (for bi-manual operation).
    \item \emph{Surgical task constraints} --- at least 2 independent curvature segments up to $\approx$90$^{\circ}$ and bending radius $<$10~mm, diameter under 5~mm (under 3~mm if navigating foramen of Monro), workspace covering $\pm$30~mm, internal cavity (diameter 1--2~mm) for optical guides and surgical instruments. 
\end{itemize}

\section{Existing Robots for Neurosurgery}

\noindent Continuum robots are a class of long and thin manipulators capable of bending along their length for extreme maneuverability and flexibility, often inspired by tentacles or elephant trunks~\cite{zhong2020recent,angrisani2019use,zhao2015surgical}. This class of device has seen medical use as catheters to reduce patient pain and for easier access to deep seated structures with natural orifices~\cite{banerjee2018soft,seah2018flexible,dupourque2019transbronchial}. However, these devices have yet to see wide reaching neurosurgical applications~\cite{zhao2015surgical,gifari2019review}. In this section, we review existing devices used in related medical applications as well as state-of-the-art potential continuum robot technologies that are still in the research phase. 

\begin{table*}[ht]
\caption{Potential Technologies for Endoscopic Neurosurgery}\label{tab:my-table}\renewcommand{\arraystretch}{1.2}
\begin{tabular}{lllllllll}
\toprule
       & Technology             & Example                                                 & Contact$^1$             & Bio-C.$^2$       & Interface$^3$  & Capabilities$^4$         & Limitations$^5$                & Fail-safes$^6$                     \\ \midrule
Rigid  & Virtual FTL            & NeoGuide~\cite{striegel2011determining}     & MR               & C                 & V            & HC, HD            & No. tendons           & EA         \\
       &                        & Tristable~\cite{calme2022towards}           & MR               & C, HT             & V            & LC, HD            & Scaling               & -          \\
       &                        & ETV-ETB TD~\cite{gao2019continuum}          & MR               & C                 & V            & MC, HD            & No. tendons           & EA         \\
 \vspace{3pt}      &                        & SMA~\cite{mandolino2022design}              & MR               & C, HT             & V            & HC, HD            & Speed                 & -          \\
       & Mechanical      & Memoslide~\cite{henselmans2017memo}         & LR               & C                 & M, A         & HC, HD            & No. tendons           & EA         \\
       & memory                       & Memobox~\cite{culmone2022memobox}           & LR               & C                 & M, A         & MC, HD            & No. tendons           & EA         \\ \midrule
Soft   & Underactuation         & Endosamurai~\cite{ikeda2011evaluation}      & HR, HF           & -                 & M            & VHC, LD           & Control               & CB         \\
 \vspace{3pt}      &                        & VNL K-flex~\cite{kim2013stiffness}          & LR               & C                 & V            & MC, HD            & No. tendons           & EA         \\
       & Growing/      & Eversion~\cite{hawkes2017soft}              & LF               & PF                & V, A         & MC, HD            & Scaling, LE           & CB         \\
       & burrowing                       & Burrowing needle~\cite{frasson2010sting}    & MR, HF           & -                 & M            & MC, LD            & Control               & -          \\ \midrule
Hybrid & Concentric  & Magnetic CTR~\cite{peyron2022magnetic}      & LR               & -                 & V, A            & LC, HD            & PR, no MRI            & -          \\
       & tube robots                       & EAP CTR~\cite{chikhaoui2018developments}    & LR               & C             & V, A            & LC, HD            & PR                    & -          \\
 \vspace{3pt}      &                        & Tendon CTR~\cite{amanov2017toward} & LR               & C                 & M, A            & LC, HD            & PR                    & EA \\
       & Deployment  & Wire jamming~\cite{kang2016first}           & LR               & C                 & M, A         & VHC, HD           & Scaling               & -          \\
       & propagation                       & Bead locking~\cite{chen2010multi}  & MR               & -                 & M, A         & VHC, HD           & SR scaling            & EA         \\ \bottomrule
\end{tabular}
\vspace{2pt}

\footnotesize{ 
$^1$HR/MR/LR --- high/moderate/low risk of high force contact when propagating. LF --- chance of low force contact. HF --- high friction.

$^2$C --- risk of contamination (no risk if disposable). HT --- high temperature. PF --- pressurised fluid.

$^3$V --- control only through virtual interface. M --- can control manually. A --- automatic follow-the-leader motion.

$^4$VHC/HC/MC/LC --- very high/high/moderate/low curvature. HD/LD --- high/low DoF (degrees of freedom).

$^5$PR --- pre-planned route required. LE --- low end-effector force. SR --- poor stiffness ratio scaling.

$^6$EA --- external actuation can be disconnected. CB --- compliant body

}
\end{table*}

Table~\ref{tab:my-table} highlights fifteen representative examples across six different primary technologies and three design paradigms: rigid, soft and hybrid robotics. Here, we compare at a high level how well each example fulfills the neurosurgical requirements outlined in Fig.~\ref{fig:2}: physical contact, bio-compatibilities/incompatibilities, user interface and capabilities towards task requirements. In addition, we list the main limitations to be overcome in each example and if the technology has any inherent fail-safes. By reviewing these examples in-depth, we can comment on the most suited design paradigm, the most promising technologies, and the challenges to overcome for minimally invasive robotic neurosurgery.

\subsection{Rigid Robotics Paradigm}

\begin{figure}[t]
    \centering
    \includegraphics[width=0.7\columnwidth]{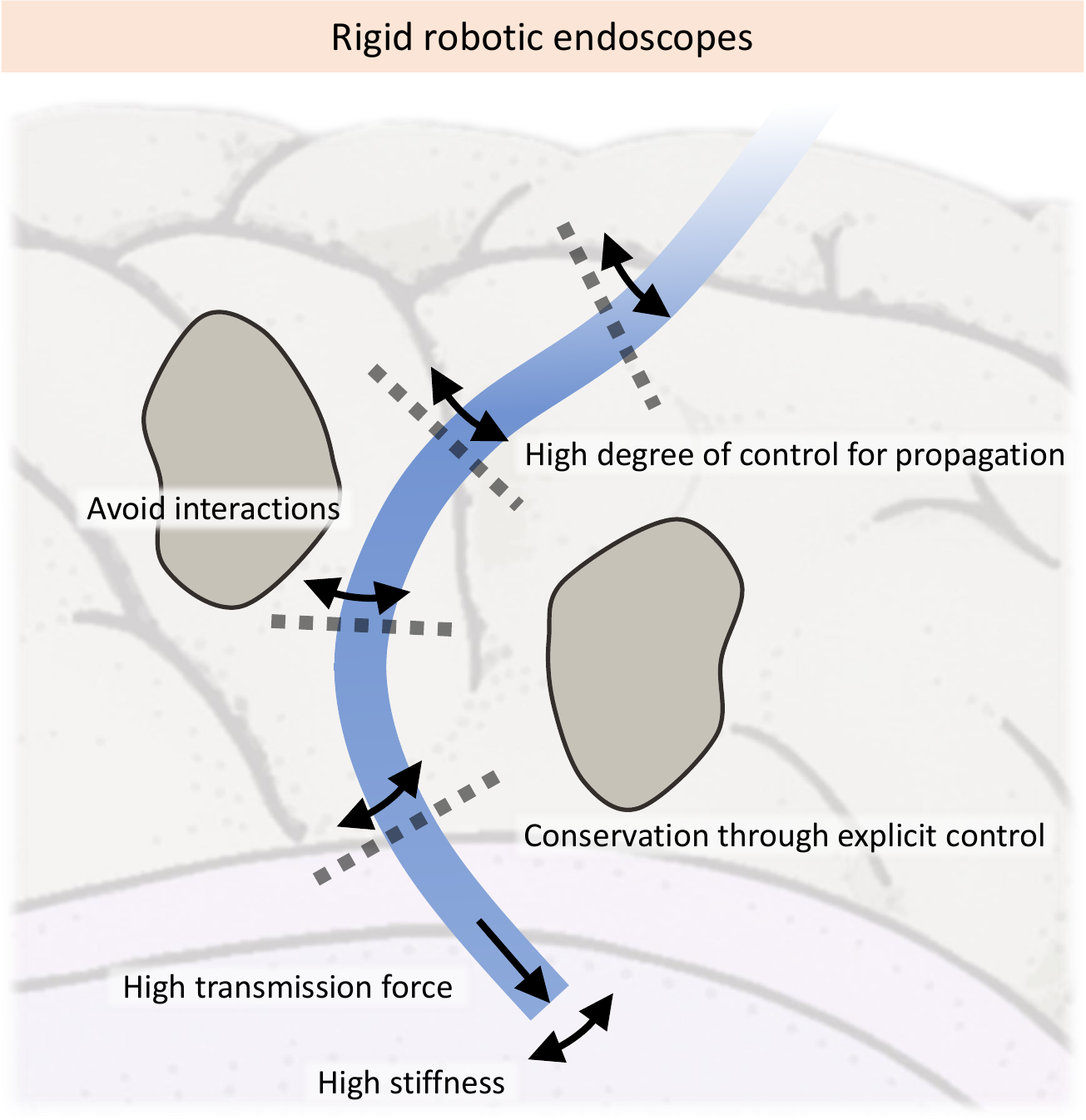}
    \caption{Rigid robots emphasise precise position control and predictability from rigid mechanisms.}
    \label{fig:4a}
\end{figure}

\noindent Rigid robots, Fig.~\ref{fig:4a} are the most traditional approach~\cite{mcbeth2004robotics,freschi2013technical,thompson2009evaluation}. Continuum robots designed around this philosophy generally maximise stiffness and independently actuate every degree of freedom in order to minimise position control error even when under external loads. Because of this, for tasks where obstacle avoidance is a priority these systems are attractive. However, the limitations of these types of systems is the complexity of actuation required and the propensity to exert large forces when contact is made with the environment~\cite{freschi2013technical}.

\begin{figure}[t]
    \centering
    \includegraphics[width=0.99\columnwidth]{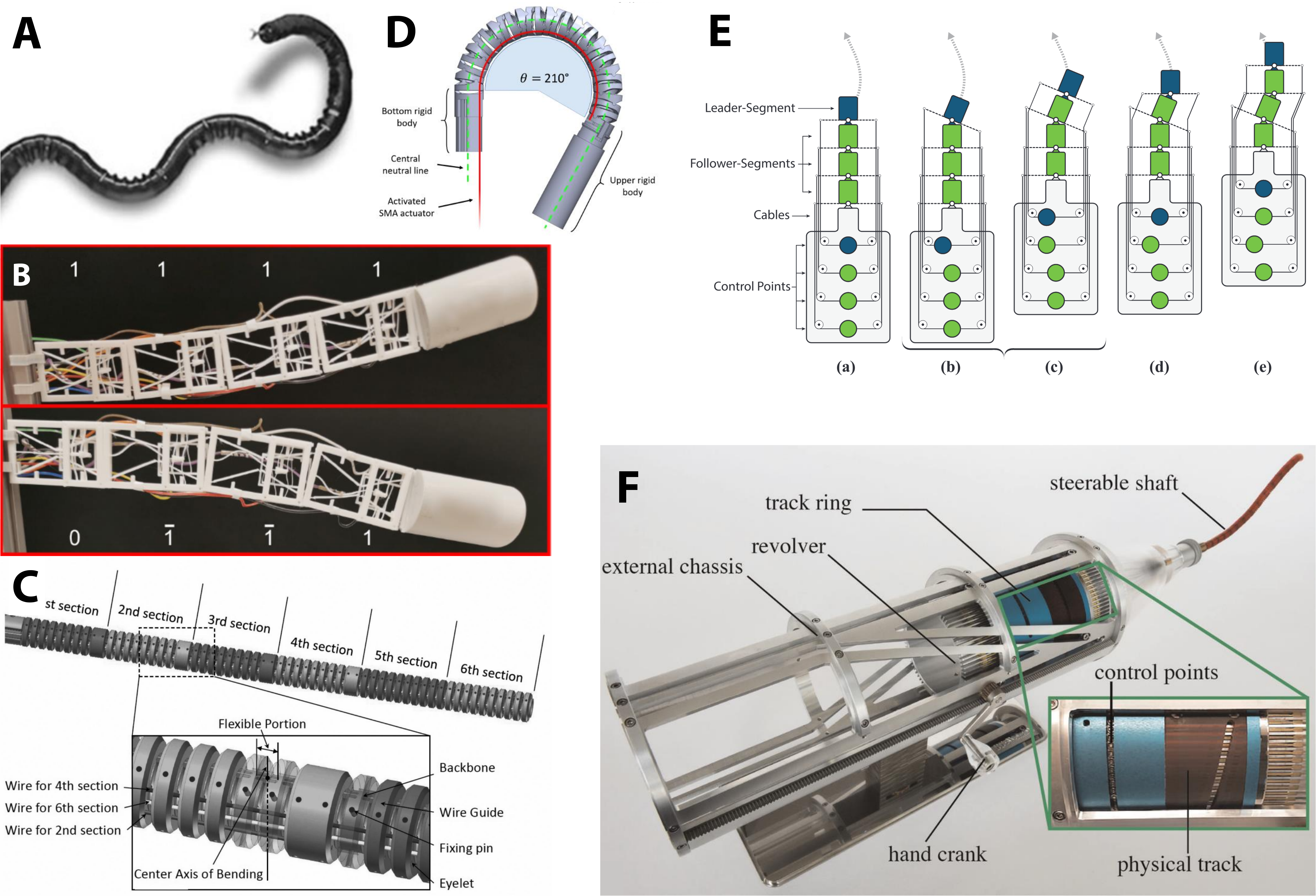}
    \caption{Rigid surgical robot examples (from Table~\ref{tab:my-table}) (A)--(F): NeoGuide~\cite{striegel2011determining}, Tristable~\protect\cite{calme2022towards}, ETV-ETB~\protect\cite{gao2019continuum}, SMA~\protect\cite{mandolino2022design}, Memoslide~\protect\cite{henselmans2017memo}, Memobox~\protect\cite{culmone2022memobox}.}
    \label{fig:4}
\end{figure}

Fig.~\ref{fig:4} shows the rigid robot representative examples from Table~\ref{tab:my-table}. The virtual follow-the-leader examples consist of independently actuated segments and, during propagation, conservation is achieved via complex control signals. Due to the discrete number of actuated segments, significant deviation can be seen from the desired path~\cite{gao2019continuum,neumann2016considerations}. This deviation is reduced with increasing number of segments, though the maximum number of segments possible is limited by any external connections such as electrical wire or tendons~\cite{culmone2021follow}. From the NeoGuide and ETV robot, we see up to 6 segments can be practical for tendon actuated systems~\cite{striegel2011determining,gao2019continuum},  Fig.~\ref{fig:4}A and C. These examples are two of the most capable, with high curvature and high degrees of freedom, giving a large workspace, and are able to be manufactured at scales under 5~mm~\cite{gao2019continuum}.

The tristable and SMA driven manipulators both move actuation internally, allowing for greater number of segments~\cite{calme2022towards} and higher curvature~\cite{mandolino2022design}. However, the tristable system has highly discretised bending capabilities, reducing its workspace, in addition to the challenge of scaling this system down to neurosurgery scale due to the mechanical and fabrication complexity. The SMA driven example offers good task capabilities, though actuation is slow and driven by joule heating of SMA wires which introduces additional risk.

The virtual FTL systems have a limited number of segments and minimum size due to space constraints with increasing number of tendons. Aside from the space constraint, each tendon must be actuated, typically with a minimum of three actuators per segment~\cite{zhao2015surgical}. Simultaneous and precise control of high degrees of freedom is expensive and a significant research challenge~\cite{zhang2022survey}.

In continuum robots designed for FTL motions, full actuation and control of every degree of freedom is unnecessary. If the bending of the manipulator can be propagated along the trunk, FTL motions can be achieved automatically without complex control and actuation. The mechanical memory examples in Table~\ref{tab:my-table} solve this with external `memory' of actuating tendon set points. The Memoslide, Fig.~\ref{fig:4}E, employs a large number of segments and a complex shifting cam mechanism to propagate tendon set points as the manipulator advances~\cite{henselmans2017memo}. FTL motions are fully automatic and have high conservation accuracy due to the large number of segments. Curvature is limited to discrete steps by the number of `memory' positions~\cite{henselmans2017memo}. The Memobox, Fig.~\ref{fig:4}F, improves upon this with 3D motion capabilities and the possibility for more continuous bending angles and manual control with a twinned controller and external tracks~\cite{culmone2022memobox,henselmans2020memoflex}. As these systems use similar manipulators with many segments and tendons, reducing the scale can be a significant challenge.

\subsection{Soft Robotics Paradigm}

\begin{figure}[t]
    \centering
    \includegraphics[width=0.7\columnwidth]{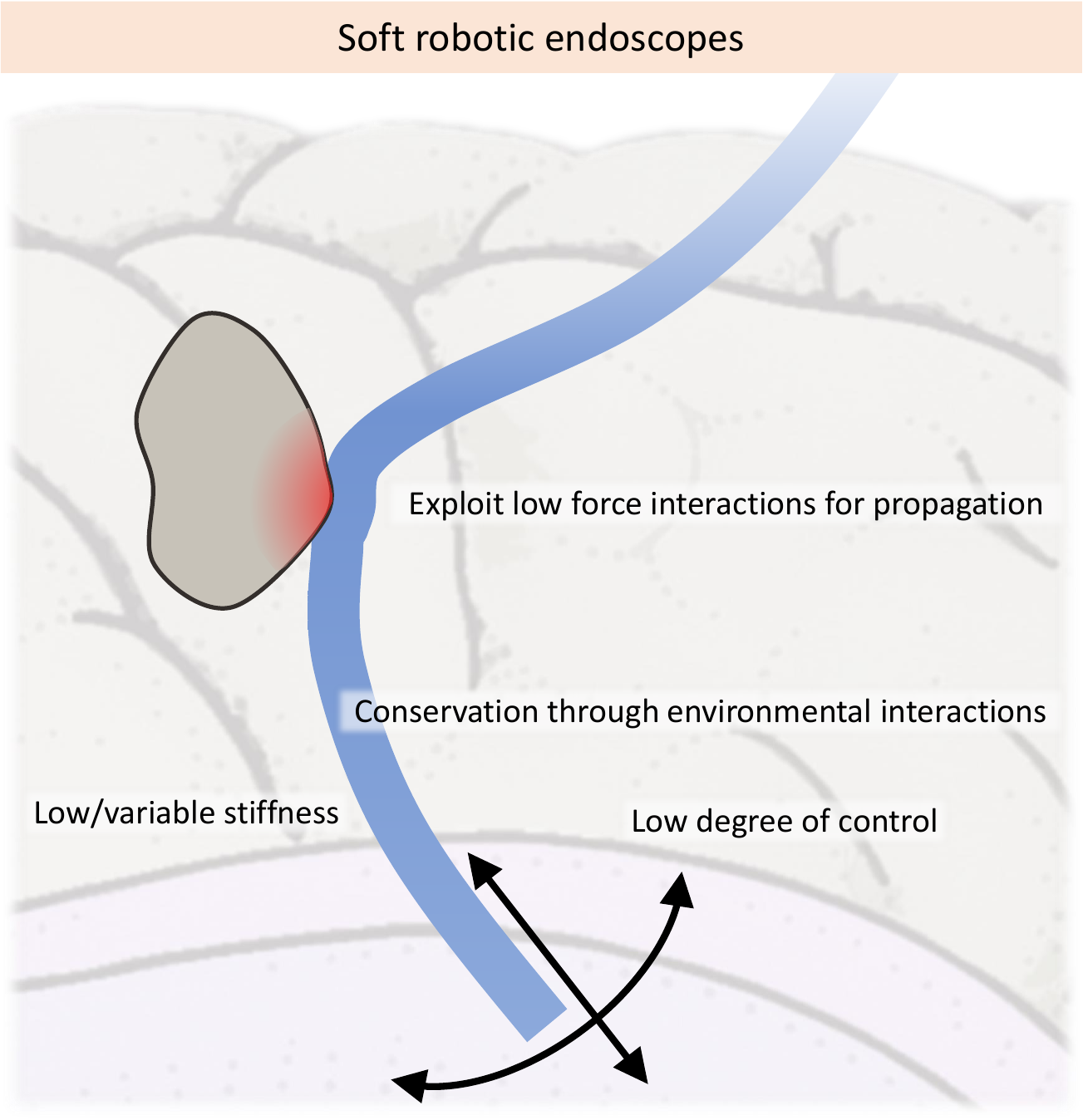}
    \caption{Soft robots emphasise intelligent material behaviours generated through environmental interactions and low degrees of actuation.}
    \label{fig:5a}
\end{figure}

\begin{figure}[t]
    \centering
    \includegraphics[width=0.99\columnwidth]{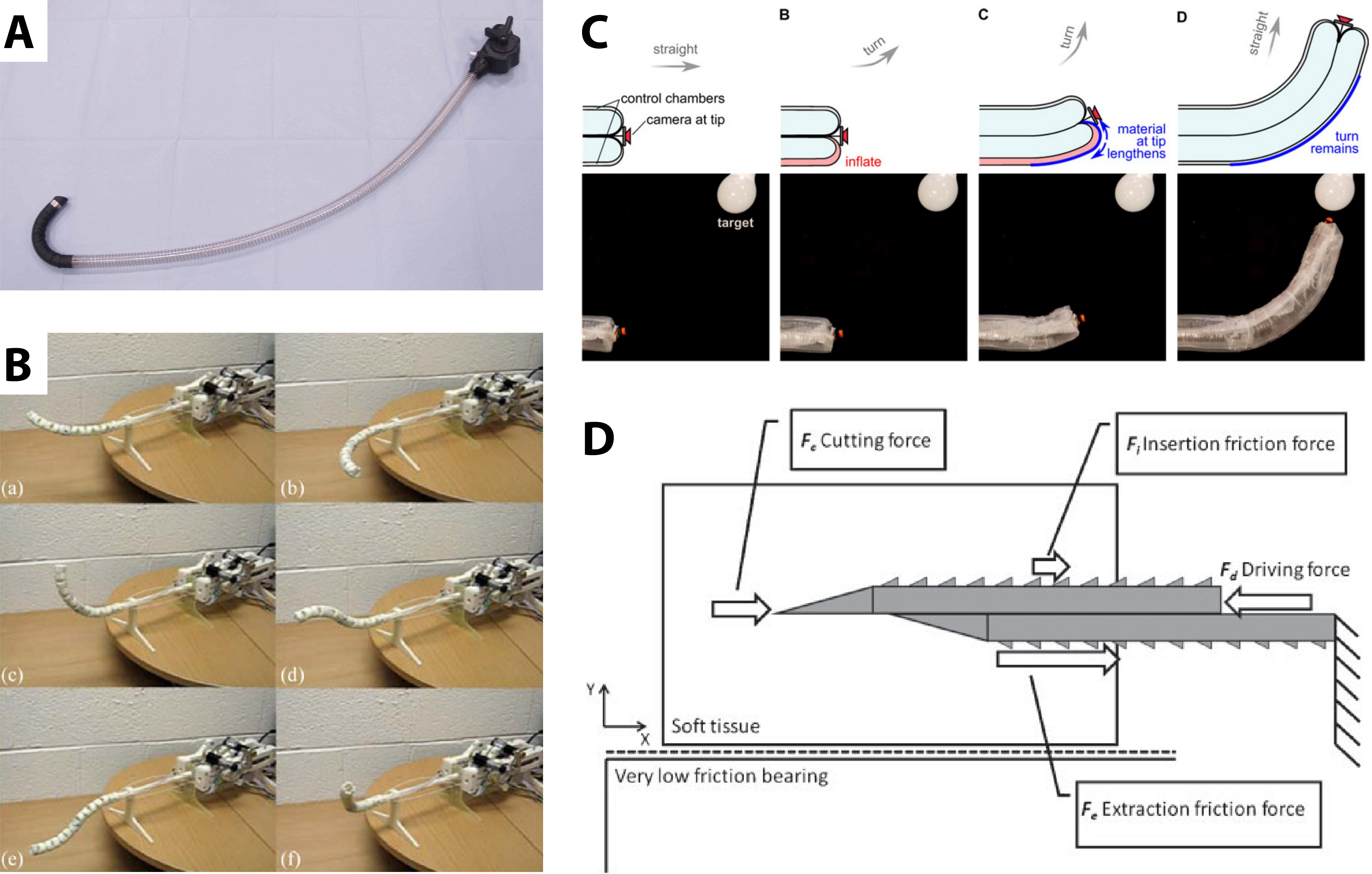}
    \caption{Soft surgical robot examples (from Table~\ref{tab:my-table}) (A)--(D): Endosamurai~\protect\cite{ikeda2011evaluation}, K-flex~\protect\cite{kim2013stiffness}, Eversion~\protect\cite{hawkes2017soft}, Burrowing~\protect\cite{frasson2010sting}.}
    \label{fig:5}
\end{figure}

\noindent Soft robots, Fig.~\ref{fig:5a}, are a relatively recent approach to the design and application of robots~\cite{gifari2019review}. In soft robotics, rigid bodies and high degrees of control are left in favour of compliant and passive-adaptive systems to generate safe environmental interactions robust to large uncertainties~\cite{angrisani2019use}. Rather than avoid interactions, these types of systems exploit interactions to generate desired responses. Flexible endoscopes illustrate this well~\cite{seah2018flexible,seeliger2020robotics}. Systems such as the Endosamurai, Fig.~\ref{fig:5}A, have essentially three controlled degrees of freedom, forward/backwards, rotation and tip angle, and FTL conformation is achieved during propagation from body interactions~\cite{ikeda2011evaluation}. This approach greatly simplifies control and mechanism design which can lead to reduced risk. However, in ventricular surgeries flexible endoscopes struggle to navigate through the cavities in more than two dimensions without touching ventricular walls, so offer little additional flexibility to reach other locations or inspect surrounding tissue. While these systems increase contact safety with inherent body compliance, there remain extremely delicate structures around the ventricles where even low force contact should be completely avoided and surgeons have to be extremely careful when advancing and retracting. Finally, the ability to complete sharp turns is limited, the sharper the turn, the greater the force needed for conformation which can lead to abrasion or even cutting of tissue during propagation~\cite{swaney2012flexure}.

Flexible endoscopes are one extreme of underactuation, keeping the same degrees of freedom but reducing the number of actuators. However, this concept can be exploited in other ways. The K-flex robot, Fig.~\ref{fig:5}B, keeps a similar number of actuators but increases the degrees of freedom, allowing compliant interactions and even control of body stiffness with antagonistic tendons~\cite{kim2013stiffness}. Similar to the fully-actuated, virtual FTL robots, without external forces, the position of the K-flex system can be controlled accurately and in high degrees of freedom with the main limitation being system complexity and the number of tendons~\cite{culmone2021follow}. While the number of tendons in this system can be reduced and flexibility can be increased by exploiting body interactions, this is not advantageous for ventricular surgeries~\cite{kahle2016hydrocephalus,gao2019continuum}.

Growing robots are another concept from the field of soft robotics~\cite{culmone2021follow}. These types of robot extend outwards from the tip, rather than being pushed through. This offers advantages in reduced friction forces and automatic FTL motions. Eversion robots demonstrate this well, Fig.~\ref{fig:5}C, where the inflation and unwrapping of a soft sleeve from the tip can allow a robot to navigate through complex spaces passively or actively with very low forces~\cite{hawkes2017soft}. Robot shape can be pre-planned or actively steered by different tip and locking mechanisms~\cite{greer2019soft}. However, this technology is still in its infancy, with many problems to solve, particularly in active steering which has limited maneuverability and often adds bulky tip modules. There are additional limitations to solve, including: the uncertainty if this technology can reach the scale required for neurosurgery; without any stiffening of the body after growth, the end effector forces are potentially too low for many surgical procedures; and inflation using a pressurised fluid introduces potentially dangerous failure modes. Similar to growing robots are burrowing robots. Burrowing needle robots, Fig.~\ref{fig:5}D, are similar to flexible endoscopes with one key difference: they pull from the tip rather than push from the base~\cite{frasson2010sting}. This overcomes the propagation force problem and can allow extreme maneuverability, however, requires contact with tissue to pull itself along~\cite{schwehr2022toward}.

\subsection{Hybrid Soft-Rigid}

\begin{figure}[t]
    \centering
    \includegraphics[width=0.7\columnwidth]{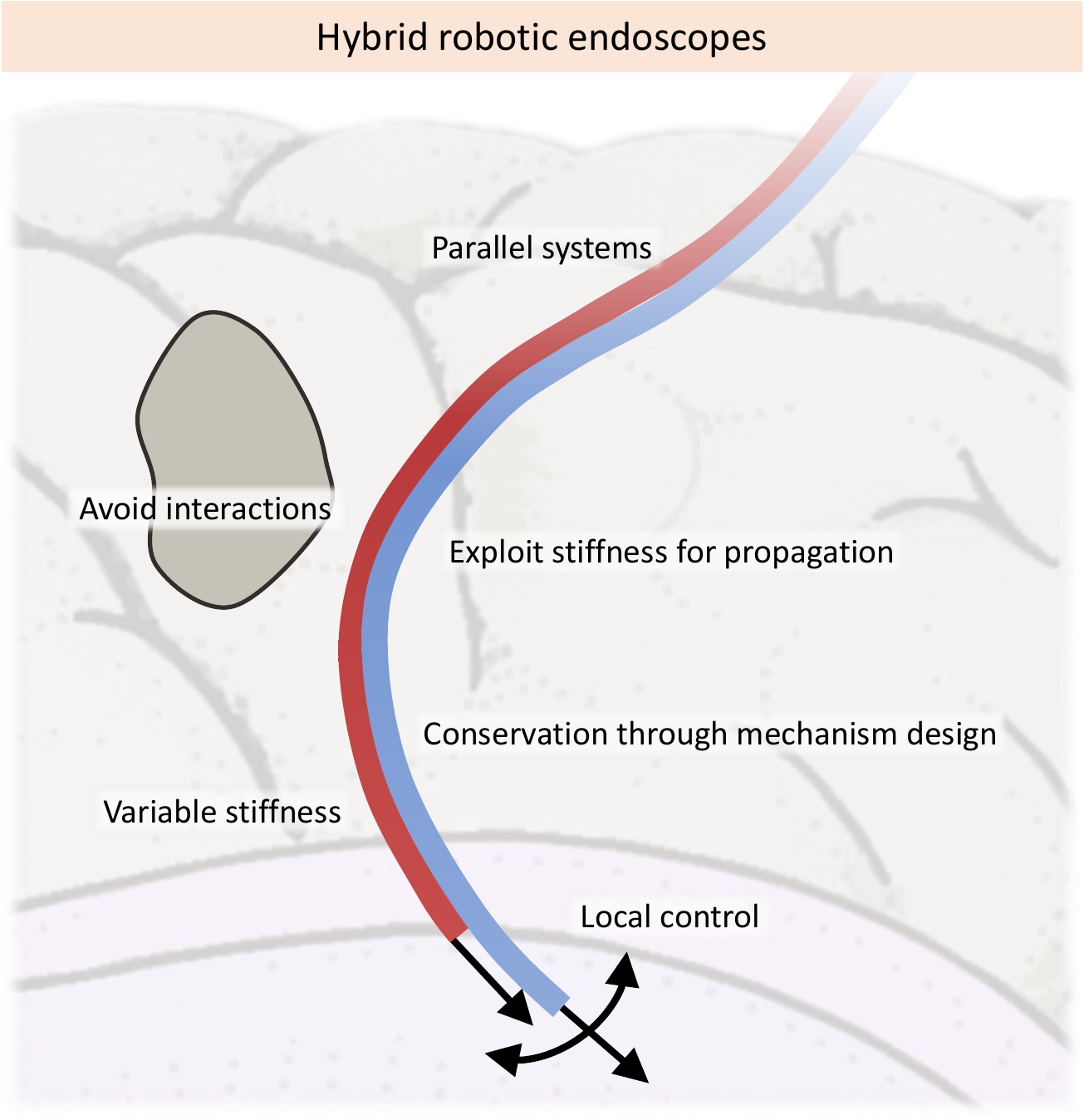}
    \caption{Hybrid robots emphasise self-interactions with variable stiffness mechanisms for high degree of freedom position control with minimal actuation and environmental interactions.}
    \label{fig:6a}
\end{figure}

\noindent Hybrid robots, Fig.~\ref{fig:6a}, attempt to combine the advantages of rigid and soft robotic systems~\cite{culha2017design,gilday2023predictive}, for example using underactuated compliant bodies along with selective stiffening for high degree of freedom control without environmental interactions~\cite{zhong2020recent}. Each of the hybrid robots in Table~\ref{tab:my-table} exploit stiffness differences within the body for propagation and conservation. Similar to the mechanical memory examples, these systems can follow FTL motions automatically with mechanical systems, though in these cases the shape `memory' to propagate along the body is stored internally in the trunk of the robot rather than externally~\cite{henselmans2017memo,culmone2022memobox}. Many examples of variable stiffness exist, the selection chosen here represent the major categories, from concentric tube robots which exploit stiffness differences in preshaped tubes by varying relative position, to active stiffness control with `jamming' and locking mechanisms~\cite{zhong2020recent}.

\begin{figure}[t]
    \centering
    \includegraphics[width=0.99\columnwidth]{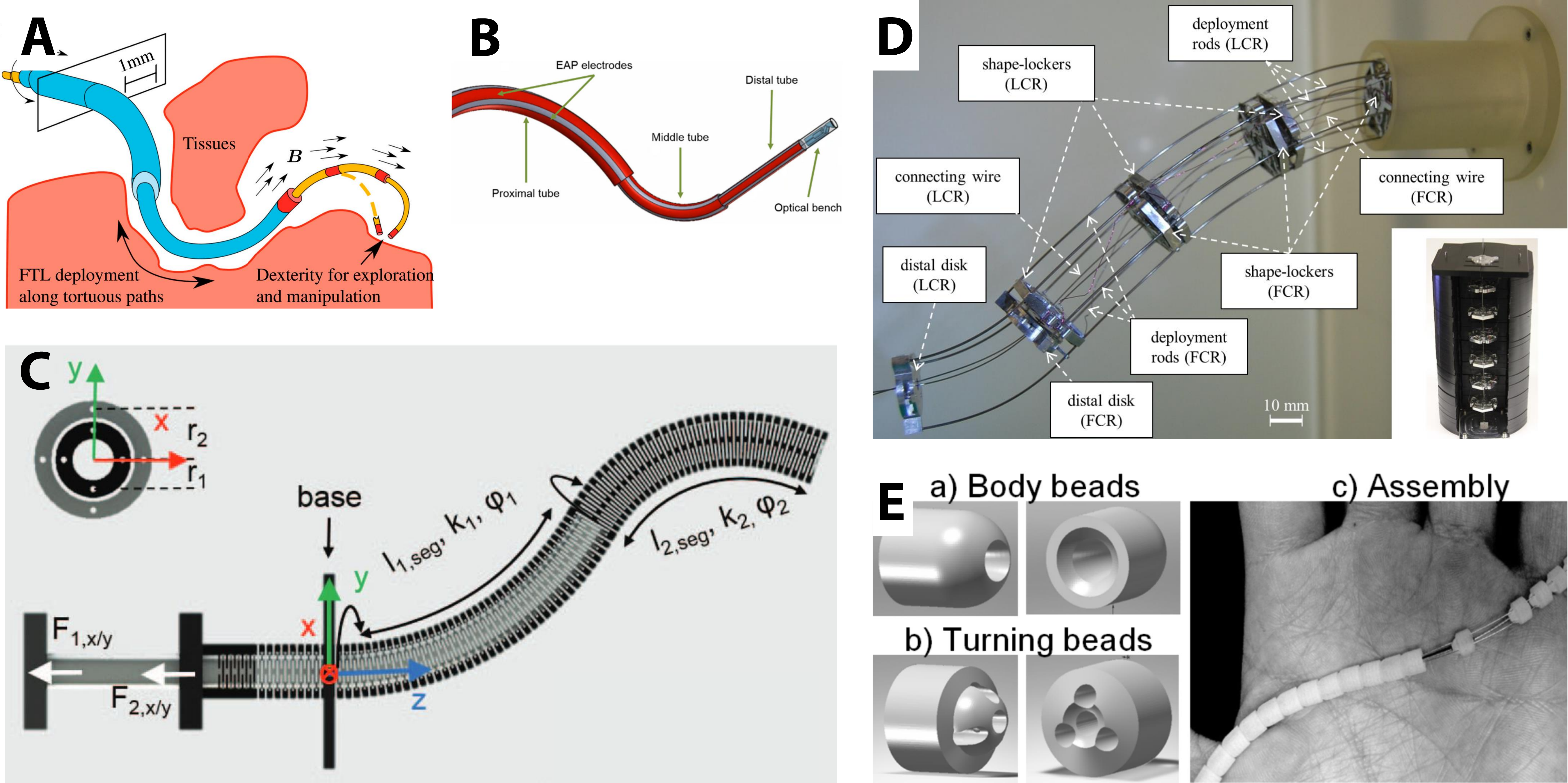}
    \caption{Hybrid surgical robot examples (from Table~\ref{tab:my-table}) (A)--(E): Magnetic CTR~\protect\cite{peyron2022magnetic}, EAP CTR~\protect\cite{chikhaoui2018developments}, Tendon CTR~\protect\cite{amanov2017toward}, Wire jamming~\protect\cite{kang2016first}, Bead locking~\protect\cite{chen2010multi}.}
    \label{fig:6}
\end{figure}

Concentric tube robots (CTRs) in particular have seen significant development and have many properties favourable for surgical applications~\cite{gilbert2015concentric}. These robots can follow complex trajectories by preforming the shape of tubes and deploying sequentially for FTL motion~\cite{garriga2018complete}. The most simple CTRs are divided into discrete segments, either purely propagating where the outer tube stiffness dominates the stiffness of the inner tubes of distal segments or with tubes of equivalent stiffness whose relative orientation defines the bending of that segment~\cite{nwafor2023design}. These robots have simple construction and need only a few actuators relative to rigid robots, additionally, they can be constructed out of easily sterilised, bio-compatible materials with very small diameters~\cite{gilbert2015concentric}.

Concentric tube robots have two major limitations for neurosurgery. The first is the limited flexibility, especially for multiple FTL routes with one set of tubes~\cite{garriga2018complete}. These robots can follow pre-planned routes with high accuracy, though in order to follow a different route or access different locations, the constituent tubes must be formed into a different shape~\cite{dupont2009design}. The second limitation is the low curvature before material elastic limits are reached~\cite{dupont2009design}. The concentric tube robots in Table~\ref{tab:my-table} are examples augmented in order to address one or both of these limitations. The magnetic CTR robot, Fig.~\ref{fig:6}A, adds a tip magnet and external magnetic field for improved control of tip position and rotation~\cite{peyron2022magnetic}. This increases the ability to access and inspect locations around a pre-planned target and demands no additional hardware or connections along the robot trunk; allowing smaller tube diameters, therefore higher curvature~\cite{culmone2021follow}. However, the routes and curvature remain limited compared with most rigid and soft designs.

The electro-active polymer (EAP) augmented CTR, Fig.~\ref{fig:6}B, attempts to increase curvature and flexibility in route following with bending actuation on each concentric tube segment~\cite{chikhaoui2018developments}. EAPs contract under applied voltage and offer low profile actuation to the tubes, increasing bending without exceeding material elastic limits. However, the EAPs themselves have limited strength, therefore currently have a minimal effect on the curvature of concentric tubes. The final CTR example, Fig.~\ref{fig:6}C, is augmented with parallel tendons to enhance bending and path following motions~\cite{amanov2017toward}. This has the potential to increase complexity with the limitations of rigid, tendon-driven manipulators, though by using telescoping concentric tubes, accurate conservation in FTL motion can be achieved with many fewer segments~\cite{amanov2017toward,wei2023coupling}. With this technology, there is a trade-off between flexibility to follow different routes (with additional segments and number of tendons) and higher curvatures and conformation (with more pre-planning, fewer and lower diameter tubes)~\cite{wu2016dexterity}.

Active stiffness control is the other major technology within the hybrid robot examples~\cite{zhong2020recent}. This tries to overcome some of the unpredictability of soft robot position control and limited end effector forces, without the actuation complexity of rigid robots, by switching between a soft conforming state and a rigid state~\cite{cianchetti2014soft}. Particle or fibre jamming are common methods for active stiffness control, though so far have limited stiffness ratios for small-scale and surgical applications. Other examples include wire jamming, Fig.~\ref{fig:6}D, which operates similar to concentric tube robots with telescoping segments~\cite{kang2016first}. Each segment, however, can be controlled in extension and bending with very high curvature, then locked in place with clamps between modules. However, the ability for this system to scale down is uncertain. 

The final example highlights the alternating stiffness control of the deployment propagation approach, Fig.~\ref{fig:6a}. In this approach, two parallel modules can be stiffened and advanced sequentially: one module is the leader, with a steerable tip, propagating forward while coupled to the stiffened follower module for FTL conservation; then the leader module stiffens and the follower module slackens and `catches up' while conserving the shape of the leader~\cite{culmone2021follow}. Fig.~\ref{fig:6}E shows the friction jamming module with a chain of beads~\cite{chen2010multi}. By increasing the tension in the tendon running through the ball and socket shaped beads, friction greatly increases, locking the beads in place. The `leader' chain of beads has a steerable tip with turning beads driven by tendons. This system offers high curvature, high flexibility in control and choice of route, and is relatively simple to actuate. The main limitation of these systems is the accuracy of control and conservation reduces with the stiffness ratio. The ratio between the stiffened and unstiffened shape reduces as the robot scales down~\cite{zhong2020recent}.

\subsection{Comparison of Approaches}

\noindent No single design fulfills all the requirements for minimally invasive robotic neurosurgery, hence the majority of intercranial procedures today are performed as open surgery or rigid endoscopic surgery~\cite{price2023using}. While some discussed designs are far from clinical trials, we are still able to evaluate the potential of each approach, whether there is a clear best solution, and what we can learn and be applied form each technology. 

In terms of safety considerations from physical contact with brain tissue, despite the philosophy of safe interactions, soft robotic approaches have some of the highest associated risks from physical contact as the brain can be preoperatively mapped and contact-free routes can be planned accordingly with little uncertainty in the environment~\cite{ferroli2013advanced,liu2021simultaneous}. In general, the model-based control of the rigid virtual FTL systems makes them well suited for tasks with high precision requirements and knowledge of posture at all times. However, due to the finite controllability of segmented manipulators and the trade-off between reduced complexity and scale versus high numbers of segments and high FTL conservation~\cite{gao2019continuum}, risk of contact cannot be fully reduced and if contact does occur it is likely to exert high forces and cause damage. Hybrid continuum manipulators are able to demonstrate FTL motion with high conformation, greatly reducing any risk of contact~\cite{peyron2022magnetic}. Additionally, the hybrid manipulators often have some in-built compliance, therefore if accidental contact occurs the force is lessened.

For the bio-compatibility of devices, the majority of reviewed technologies can be fabricated from inert materials such as plastics or medical grade alloys such as nitinol~\cite{peyron2022magnetic}. No major incompatibilities are seen, with some minor hazards that can be mitigated, such as heat elements, internal voltages or operating fluids/particles~\cite{cianchetti2014soft}. Otherwise the main concern is device sterilisation in the case of reuse. Ideally the device should be reusable, the most complex and least desirable as single-use tools are also the most challenging to sterilise with many nooks and overlapping structures~\cite{amanov2017toward}. This is most prevalent in the highly articulated rigid robot manipulators~\cite{gao2019continuum}.

The surgeon-device interface is linked to the safety and capabilities of the device. Complex or unintuitive devices must be operated through a virtual interface~\cite{striegel2011determining}. This interface can generate control signals needed for FTL motions in rigid robots, control stiffness in soft robots, or provide the necessary coordination and parallel system mapping in hybrid robots~\cite{gao2019continuum,kim2013stiffness,peyron2022magnetic}. Virtual interfaces are often necessary in microsurgery for the highest precision, though one critical drawback is the lack of haptic feedback and the lack of trust in more autonomous systems~\cite{zhao2015surgical,elsabeh2021cranial}. If a device can be controlled directly with a manual interface, the surgeon can receive direct force feedback. Rigid robots have the capability for manual control if the FTL motion is generated mechanically rather than virtually~\cite{henselmans2017memo}, this reduces the degrees of control down to steering and propagation which is highly intuitive. Soft robots vary in their capabilities here, where some can be controlled manually, but not necessarily with FTL motions (flexible endoscopes~\cite{ikeda2011evaluation,seah2018flexible}), some require a virtual interface similar to rigid robots and some require virtual interfaces even with automatic FTL motion as actuation requires non-manual processes such as pressure control with eversion robots~\cite{hawkes2017soft}. Hybrid systems in contrast all can generate automatic FTL motions and most can be driven manually~\cite{zhong2020recent}. There are exceptions, high number of concentric tubes can be difficult to control accurately by hand, deployment propagation systems can be unintuitive to control, and actuation from external magnetic fields or actuators embedded in the body of the manipulator must be controlled virtually~\cite{peyron2022magnetic,chikhaoui2018developments}.

The extent to which technology is suited for a given surgery depends on the requirements of the surgical task. The specification can include robot scale, working envelope, end-effector carrying capacity and more. Here we just consider how capable a robot is to reach arbitrary locations. Devices with high curvature relative to their length are needed for sharp turns within the ventricles and high degrees of freedom are essential for following complex paths either to hard to reach places or accessing sites from different angles~\cite{anderson2003surgical,price2023using}. Rigid robots excel in degrees of freedom and can navigate easily on arbitrary routes, their maximum curvature can be limited especially when the bending is achieved with a compliant backbone rather than a more complex articulated system. Soft robots once again have a range of capabilities, with the highest performing examples being the variable stiffness K-flex robot~\cite{kim2013stiffness} and the eversion robot~\cite{hawkes2017soft}, maximum curvature can potentially be improved by adjusting design parameters. Hybrid robots can generally follow arbitrary routes with high degrees of freedom, though CTRs with bending dependent on tube elasticity have low curvature relative to their length before exceeding material limits~\cite{garriga2018complete}.

Rigid and soft robots have similar limitations with miniaturization and the number of tendons required. Other limitations which could reduce applicability in some surgical tasks are low end-effector force capabilities in soft robots when interacting with surgical targets, one reason why variable stiffness designs are often preferred~\cite{gifari2019review,chen2010multi}. Control can be another limitation, either highly restrictive in CTRs with FTL capabilities limited to pre-planned routes, or too unpredictable, especially within natural cavities, such as with flexible endoscopes or burrowing needles~\cite{frasson2010sting}. These limitations stand out as major technical problems that need to be solved before clinical trials.

\subsection{Summary}

\noindent In their current state, rigid robots are the most suited for ventricular neurosurgeries. These robots have had significant development historically~\cite{mcbeth2004robotics,faria2015review,zhao2015surgical} and generally satisfy the requirements for neurosurgery even if not excelling in any specific requirement. Tendon-driven designs appear the most common, simplifying design and allowing bulky actuation to be placed externally. The manipulators for virtual FTL systems and mechanical memory systems are similar, with the main difference being the control mechanism. Fully actuated virtual FTL designs are more common, but do not necessarily have significant advantages over mechanical memory designs. While challenges in actuation and fabrication complexity can be overcome, it remains desirable to simplify systems --- the fewer components, actuators and mechanisms present, the fewer potential points of failure.

Despite the success of flexible endoscopes in other surgical tasks, soft robots are generally ill-suited in neurosurgical tasks with their primary advantages in flexibility and exploiting environmental interactions being potentially harmful in delicate operations. There are important conclusions we can see from the development of soft neurosurgical robots. For one, their influence on other designs to increase interaction safety~\cite{banerjee2018soft,gifari2019review}. Soft robotics exploits material properties to generate desired behaviours, in this way, complex mechanisms can be simplified, for example by `storing' position memory in the physical body with the deployment propagation robots~\cite{chen2010multi}. We observe the K-flex systems fulfills the neurosurgical requirements the best out of all the soft robots reviewed. This design is closest to the rigid robots, but takes advantage of an underactuated compliant manipulator with non-linear antagonistic tendon control for variable stiffness, with a cost of small increase in actuation complexity~\cite{kim2013stiffness}.

Hybrid robotic endoscopes attempt to take the advantages of control predictability of rigid robots and the simplicity of soft robots into new designs. This is demonstrated to some success with impressive automatic FTL capabilities, with conservation even exceeding many highly segmented rigid robot designs~\cite{gilbert2015concentric}, and intuitive control with few actuators. The best FTL capabilities are seen with the CTRs, which are well suited to neurosurgery across multiple categories, though by themselves, lack the flexibility for more complex surgeries and their limited curvature further restricts applications. Deployment propagation promises to solve all of the above problems. However, these systems are prone to poor scaling, especially when relying on friction effects to vary stiffness~\cite{zhong2020recent,chen2010multi}. As these devices decrease in scale, their variable stiffness ratio reduces and controllability worsens. While these technologies show disadvantages compared with current rigid robots, they offer interesting and varied designs which have the potential to outperform existing rigid robots with some technical challenges solved. Additionally, new hybrid designs are constantly emerging; instead of iterating on established technologies, the field is open to step changes in performance.

\section{Remaining Challenges}

\noindent The current limitations of flexible neurosurgical robots means safety cannot be guaranteed. In order to offer improved patient outcomes compared to existing open and rigid endoscopic approaches, we have identified a number of challenges to solve. Not only are there technological challenges around fulfilling the requirements for minimally invasive neurosurgery, there are challenges around operating procedures, device usability, and device evaluation and training.

\subsection{Rigid/Soft/Hybrid Robot Technology}

\noindent Rigid style continuum manipulators could be used for neurosurgical applications if there is an incremental improvement in curvature and scale. The more pressing technological challenge with these is the guarantee of safety, especially as they have the highest number of moving parts and points of failure out of all the devices reviewed here. This requires extensive testing and possibly the development of additional monitoring systems such as medical imaging based position feedback~\cite{doulgeris2015robotics,faria2015review}. 

In soft robotic devices, materials and fabrication are the primary challenges. There is much to explore with different polymers and advanced smart materials that allow us to program complex robot behaviours directly into its structure~\cite{pishvar2020foundations}. Related to this is fabrication, these robots are often made from 3D printing or casting. The scale of design with the fabrication techniques available for our desired soft materials is too large for neurosurgical applications~\cite{momeni2017review}.

Hybrid and soft robotics offers numerous actuation methods and other mechanism designs for propagation, steering and conservation of continuum manipulators~\cite{zhong2020recent}. This opens up possibilities for different combinations, series and parallel mechanisms and wholly new designs. For example, integrating a layer or particle jamming mechanism into eversion robots can offer a compact solution to the limited force capabilities and provide better guarantee of contact avoidance~\cite{do2020dynamically}. Exploration of this design space is a challenge, requiring inventiveness and exploitation of new materials and fabrication techniques.

\subsection{Haptic feedback}

\noindent Virtual/teleoperated surgical robots isolate the surgeon from the patient. With no force or proprioceptive feedback surgeons are more likely to cause tissue damage~\cite{wagner2002role}. With increasingly complex manipulators, unintuitive robot mechanics, remote and highly precise microsurgery, virtual interfaces may be necessary for many applications and haptic feedback offers a method of reconnecting the surgeon and patient~\cite{van2009value}. Indeed, haptic feedback may be preferable to direct feedback in some cases, particularly in low force applications where the haptic interface can amplify measured forces or transform coordinate systems so the force feedback matches the orientation of the end-effector visual feedback.

Force feedback can be provided from internal or external sensing elements. Strain gauges are one example, though for more compliant robots soft sensors are necessary~\cite{watanabe2014force}. Sensing in endoscopic devices faces challenges, including miniaturisation, connecting wires and sterilisation~\cite{doulgeris2015robotics}. Sensors placed in the tip of the robot yield the most precise force feedback, though compete for space with end-effector tools and the continuum manipulator actuation. This is one area where `smart' materials could provide benefits, with sensors embedded in the material itself~\cite{pishvar2020foundations}.

\subsection{Evaluation \& Testing Methodology}

\begin{figure}[t]
    \centering
    \includegraphics[width=0.99\columnwidth]{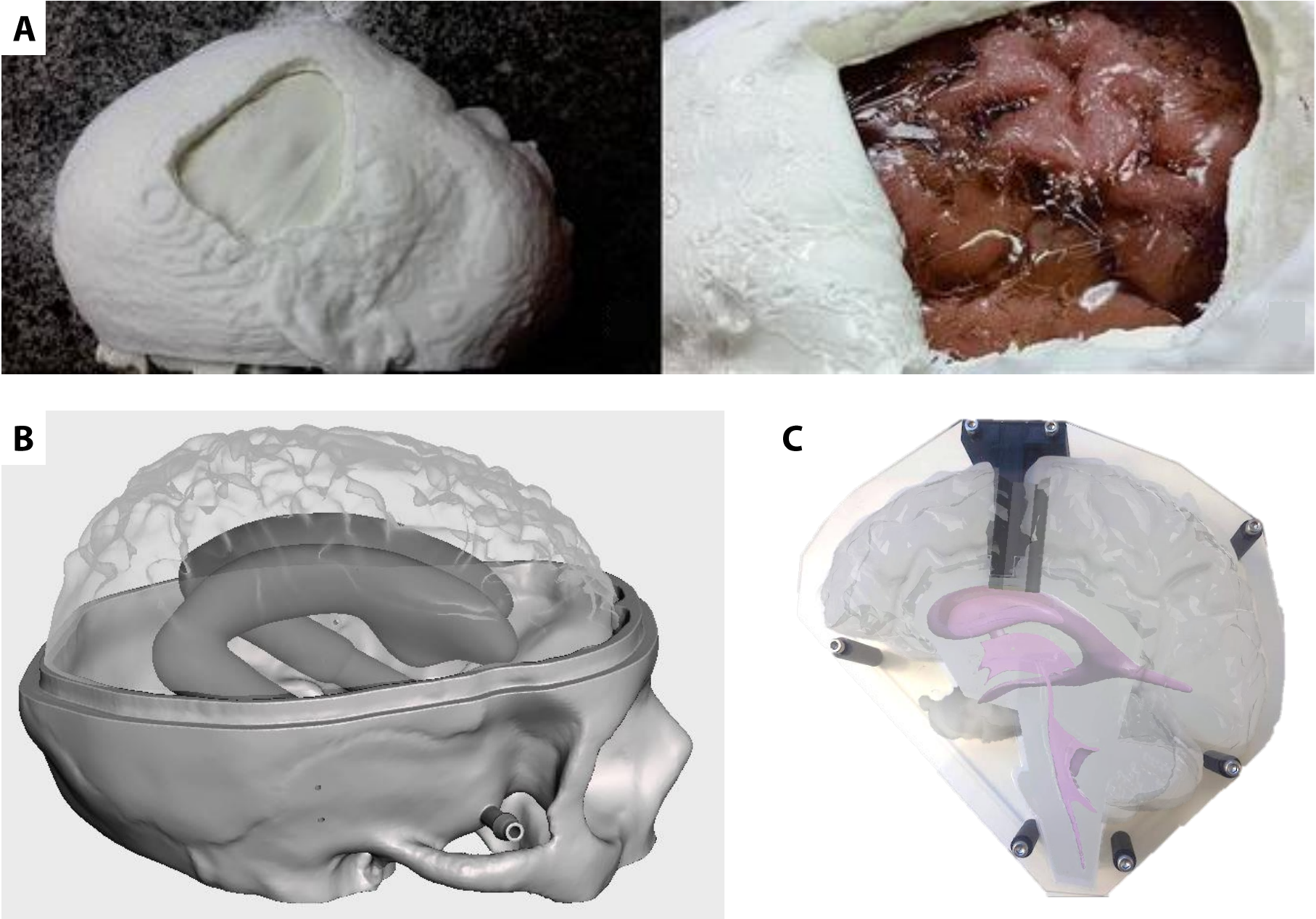}
    \caption{Evaluation techniques for robotic surgical phantoms, (A)--(C): MRI prototyped brain~\protect\cite{grillo2018patient}, Viomerse surgical simulator, Ventricular system phantom for surgical force feedback.}
    \label{fig:7}
\end{figure}

\noindent Development of flexible robotic tools is often performed in isolation from real-world applications or feedback from surgeons. Clinical trials and consultations with end users are critical for any medical device, though a more closely coupled development process would greatly benefit both roboticists and surgeons. One method for this is with more readily available neurosurgical simulations and brain phantoms, especially those that provide useful feedback on surgical outcomes~\cite{collins1998design,altermatt2019design,breimer2015design}. A number of commercial phantoms are available, these are primarily used for training and teaching brain structures. Fig.~\ref{fig:7}A shows a rapidly prototyped brain reconstructed from MRI with realistic physical properties for training in medical education~\cite{grillo2018patient}. Viomerse, Fig.~\ref{fig:7}B, offers commercial neurosurgical phantoms with some procedural feedback such as blood loss, tumor margins and operation time. Fig.~\ref{fig:7}C shows the ventricular system in a soft phantom with future potential for integrated pressure sensing. For development of robotic tools, a sensorised phantom could provide useful feedback in the form of contact detection, interaction force and ease of device use.

Related to the development of evaluation tools for robot development is a more quantitative specification of neurosurgical requirements. In this work, we are able to compare devices with each other, but since there is no objective measurement of acceptable pressures before damage mapped over the ventricular system, robot design is constrained to those that avoid contact at all costs. Even minor differences between patients brains can change the particular surgical specifications, for this reason and the uncertainty in operating in such critical spaces surgical robots must take a more general-purpose approach~\cite{price2023using}. If boundaries are more clearly defined, such as contact pressures, curvatures and degrees of freedom, bespoke devices can be created with appropriate safety factors and reduce risk.

\subsection{Future Outlook}

\noindent Overall, there is a positive outlook for the future of robotic neurosurgery. Robots are well situated to address the shortcomings of current neurosurgical procedures. There remains a number of challenges to increase precision and reduce invasiveness. Additionally, there are many opportunities in the development of novel technologies to address these problems in new ways and in providing meaningful feedback on surgical performance. 

Due to the necessary standards and regulation surrounding medical devices, development and real-world use is delayed~\cite{banerjee2018soft}. Guarantees of safety are only possible with rigorous testing and analysis. This could be one reason soft and hybrid robots have to date seen little success as commercial neurosurgical devices and over the next 5--10 years we may see products emerge based-on some of the technologies reviewed in this paper, such as mechanical memory, underactuation or more flexible concentric tubes. 

While technological progress in design and modelling of structures can increase our confidence in devices, the real-world introduces defects, unexpected situations and human-based errors such that it is impossible to account for everything at the design stage. Brain phantoms and clinical trials can go some way towards testing under a variety of conditions and environments. Though, looking to the future, taking advantage of large-scale data such as from automated robotic testing and phantom feedback can allow for rapid evaluation over diverse conditions.

\section{Conclusions}

\noindent In conclusion, current neurosurgical robot designs have limited flexibility and guarantee of safety required for improved patient outcomes. Traditional approaches focus on rigid body mechanics and precise control over many degrees of freedom and complex models to avoid contact wherever possible. Soft robotics seeks to exploit low force contact and tissue interactions for increased flexibility, adaptive behaviours and reduced control complexity. Hybrid robots attempt to combine the increased flexibility and low contact forces of soft robots with the precise control and predictability of rigid robots, while generating automatic follow-the-leader motions. We see some success towards more capable neurosurgical robots in each of these areas, with key lessons of: reduced system complexity for fewer points of failure and better capabilities to provide force-feedback to the surgeon; soft interactions can do more harm than good with highly critical and delicate structures within regions such as the ventricular system; and exploitation of variable stiffness interactions rather than environmental interactions is advantageous for safe, controlled motions inside brain cavities, though considerations must be made for curvature and force scaling during miniaturisation.

We observe that rigid type robots have seen the most progress towards neurosurgical operations, though the innovative hybrid robot designs show promise for future development and open opportunities for novel mechanisms towards automatic follow-the-leader motions, high curvature and small scale, meanwhile providing low force contact, bio-compatibility and intuitive user control. Other than challenges in the design of such systems, we have identified materials and fabrication techniques as major challenges, in addition to systems for device evaluation, haptic feedback interfaces and how to guarantee safety within devices to fulfill ethical and regulatory obligations.

\section*{Acknowledgements}

This work was supported by the Zeiss/EPFL Research-IDEAS initiative. We thank Markus Philipp, Ernar Amanov and Arvind N. Vadivelu for their support during this project.

\bibliographystyle{IEEEtran}
\bibliography{IEEEabrv,refs.bib}

\begin{thebibliography}{10}
\providecommand{\url}[1]{#1}
\csname url@samestyle\endcsname
\providecommand{\newblock}{\relax}
\providecommand{\bibinfo}[2]{#2}
\providecommand{\BIBentrySTDinterwordspacing}{\spaceskip=0pt\relax}
\providecommand{\BIBentryALTinterwordstretchfactor}{4}
\providecommand{\BIBentryALTinterwordspacing}{\spaceskip=\fontdimen2\font plus
\BIBentryALTinterwordstretchfactor\fontdimen3\font minus \fontdimen4\font\relax}
\providecommand{\BIBforeignlanguage}[2]{{%
\expandafter\ifx\csname l@#1\endcsname\relax
\typeout{** WARNING: IEEEtran.bst: No hyphenation pattern has been}%
\typeout{** loaded for the language `#1'. Using the pattern for}%
\typeout{** the default language instead.}%
\else
\language=\csname l@#1\endcsname
\fi
#2}}
\providecommand{\BIBdecl}{\relax}
\BIBdecl

\bibitem{elsabeh2021cranial}
R.~Elsabeh, S.~Singh, J.~Shasho, Y.~Saltzman, and J.~M. Abrahams, ``Cranial neurosurgical robotics,'' \emph{British journal of neurosurgery}, vol.~35, no.~5, pp. 532--540, 2021.

\bibitem{price2023using}
K.~Price, J.~Peine, M.~Mencattelli, Y.~Chitalia, D.~Pu, T.~Looi, S.~Stone, J.~Drake, and P.~E. Dupont, ``Using robotics to move a neurosurgeon’s hands to the tip of their endoscope,'' \emph{Science Robotics}, vol.~8, no.~82, p. eadg6042, 2023.

\bibitem{zhao2015surgical}
J.~Zhao, B.~Feng, M.-H. Zheng, and K.~Xu, ``Surgical robots for spl and notes: A review,'' \emph{Minimally Invasive Therapy \& Allied Technologies}, vol.~24, no.~1, pp. 8--17, 2015.

\bibitem{li2005neuroendoscopy}
K.~W. Li, C.~Nelson, I.~Suk, and G.~I. Jallo, ``Neuroendoscopy: past, present, and future,'' \emph{Neurosurgical focus}, vol.~19, no.~6, pp. 1--5, 2005.

\bibitem{demerdash2017endoscopic}
A.~Demerdash, B.~G. Rocque, J.~Johnston, C.~J. Rozzelle, B.~Yalcin, R.~Oskouian, J.~Delashaw, and R.~S. Tubbs, ``Endoscopic third ventriculostomy: A historical review,'' \emph{British journal of neurosurgery}, vol.~31, no.~1, pp. 28--32, 2017.

\bibitem{nathoo2005touch}
N.~Nathoo, M.~C. {\c{C}}avu{\c{s}}o{\u{g}}lu, M.~A. Vogelbaum, and G.~H. Barnett, ``In touch with robotics: neurosurgery for the future,'' \emph{Neurosurgery}, vol.~56, no.~3, pp. 421--433, 2005.

\bibitem{mcbeth2004robotics}
P.~B. McBeth, D.~F. Louw, P.~R. Rizun, and G.~R. Sutherland, ``Robotics in neurosurgery,'' \emph{The American Journal of Surgery}, vol. 188, no.~4, pp. 68--75, 2004.

\bibitem{faria2015review}
C.~Faria, W.~Erlhagen, M.~Rito, E.~De~Momi, G.~Ferrigno, and E.~Bicho, ``Review of robotic technology for stereotactic neurosurgery,'' \emph{IEEE reviews in biomedical engineering}, vol.~8, pp. 125--137, 2015.

\bibitem{zhong2020recent}
Y.~Zhong, L.~Hu, and Y.~Xu, ``Recent advances in design and actuation of continuum robots for medical applications,'' in \emph{Actuators}, vol.~9, no.~4.\hskip 1em plus 0.5em minus 0.4em\relax MDPI, 2020, p. 142.

\bibitem{gifari2019review}
M.~W. Gifari, H.~Naghibi, S.~Stramigioli, and M.~Abayazid, ``A review on recent advances in soft surgical robots for endoscopic applications,'' \emph{The International Journal of Medical Robotics and Computer Assisted Surgery}, vol.~15, no.~5, p. e2010, 2019.

\bibitem{rhoton2002lateral}
A.~L. Rhoton~Jr, ``The lateral and third ventricles,'' \emph{Neurosurgery}, vol.~51, no.~4, pp. S1--207, 2002.

\bibitem{angrisani2019use}
L.~Angrisani, S.~Grazioso, G.~Di~Gironimo, D.~Panariello, and A.~Tedesco, ``On the use of soft continuum robots for remote measurement tasks in constrained environments: A brief overview of applications,'' in \emph{2019 IEEE International Symposium on Measurements \& Networking (M\&N)}.\hskip 1em plus 0.5em minus 0.4em\relax IEEE, 2019, pp. 1--5.

\bibitem{banerjee2018soft}
H.~Banerjee, Z.~T.~H. Tse, and H.~Ren, ``Soft robotics with compliance and adaptation for biomedical applications and forthcoming challenges,'' \emph{Int. J. Robot. Autom}, vol.~33, no.~1, pp. 68--80, 2018.

\bibitem{seah2018flexible}
T.~E.~T. Seah, T.~N. Do, N.~Takeshita, K.~Y. Ho, and S.~J. Phee, ``Flexible robotic endoscopy systems and the future ahead,'' \emph{Diagnostic and Therapeutic Procedures in Gastroenterology: An Illustrated Guide}, pp. 521--536, 2018.

\bibitem{kume2016flexible}
K.~Kume, ``Flexible robotic endoscopy: current and original devices,'' \emph{Computer Assisted Surgery}, vol.~21, no.~1, pp. 150--159, 2016.

\bibitem{doulgeris2015robotics}
J.~J. Doulgeris, S.~A. Gonzalez-Blohm, A.~K. Filis, T.~M. Shea, K.~Aghayev, and F.~D. Vrionis, ``Robotics in neurosurgery: evolution, current challenges, and compromises,'' \emph{Cancer Control}, vol.~22, no.~3, pp. 352--359, 2015.

\bibitem{culmone2021follow}
C.~Culmone, S.~F. Yikilmaz, F.~Trauzettel, and P.~Breedveld, ``Follow-the-leader mechanisms in medical devices: A review on scientific and patent literature,'' \emph{IEEE Reviews in Biomedical Engineering}, 2021.

\bibitem{siepel2021needle}
F.~J. Siepel, B.~Maris, M.~K. Welleweerd, V.~Groenhuis, P.~Fiorini, and S.~Stramigioli, ``Needle and biopsy robots: A review,'' \emph{Current Robotics Reports}, vol.~2, pp. 73--84, 2021.

\bibitem{kahle2016hydrocephalus}
K.~T. Kahle, A.~V. Kulkarni, D.~D. Limbrick, and B.~C. Warf, ``Hydrocephalus in children,'' \emph{The lancet}, vol. 387, no. 10020, pp. 788--799, 2016.

\bibitem{anderson2003surgical}
R.~C. Anderson, S.~Ghatan, and N.~A. Feldstein, ``Surgical approaches to tumors of the lateral ventricle,'' \emph{Neurosurgery Clinics}, vol.~14, no.~4, pp. 509--525, 2003.

\bibitem{abbassy2018outcome}
M.~Abbassy, K.~Aref, A.~Farhoud, and A.~Hekal, ``Outcome of single-trajectory rigid endoscopic third ventriculostomy and biopsy in the management algorithm of pineal region tumors: a case series and review of the literature,'' \emph{Child's Nervous System}, vol.~34, pp. 1335--1344, 2018.

\bibitem{gao2019continuum}
Y.~Gao, K.~Takagi, T.~Kato, N.~Shono, and N.~Hata, ``Continuum robot with follow-the-leader motion for endoscopic third ventriculostomy and tumor biopsy,'' \emph{IEEE Transactions on Biomedical Engineering}, vol.~67, no.~2, pp. 379--390, 2019.

\bibitem{henselmans2017memo}
P.~W. Henselmans, S.~Gottenbos, G.~Smit, and P.~Breedveld, ``The memo slide: An explorative study into a novel mechanical follow-the-leader mechanism,'' \emph{Proceedings of the Institution of Mechanical Engineers, Part H: Journal of Engineering in Medicine}, vol. 231, no.~12, pp. 1213--1223, 2017.

\bibitem{ikeda2011evaluation}
K.~Ikeda, K.~Sumiyama, H.~Tajiri, K.~Yasuda, and S.~Kitano, ``Evaluation of a new multitasking platform for endoscopic full-thickness resection,'' \emph{Gastrointestinal endoscopy}, vol.~73, no.~1, pp. 117--122, 2011.

\bibitem{peyron2022magnetic}
Q.~Peyron, Q.~Boehler, P.~Rougeot, P.~Roux, B.~J. Nelson, N.~Andreff, K.~Rabenorosoa, and P.~Renaud, ``Magnetic concentric tube robots: introduction and analysis,'' \emph{The International Journal of Robotics Research}, vol.~41, no.~4, pp. 418--440, 2022.

\bibitem{hawkes2017soft}
E.~W. Hawkes, L.~H. Blumenschein, J.~D. Greer, and A.~M. Okamura, ``A soft robot that navigates its environment through growth,'' \emph{Science Robotics}, vol.~2, no.~8, p. eaan3028, 2017.

\bibitem{striegel2011determining}
J.~Striegel, R.~Jakobs, J.~Van~Dam, U.~Weickert, J.~F. Riemann, and A.~Eickhoff, ``Determining scope position during colonoscopy without use of ionizing radiation or magnetic imaging: the enhanced mapping ability of the neoguide endoscopy system,'' \emph{Surgical endoscopy}, vol.~25, pp. 636--640, 2011.

\bibitem{culmone2022memobox}
C.~Culmone, D.~J. Jager, and P.~Breedveld, ``Memobox: A mechanical follow-the-leader system for minimally invasive surgery,'' \emph{Frontiers in Medical Technology}, vol.~4, p. 938643, 2022.

\bibitem{kang2016first}
B.~Kang, R.~Kojcev, and E.~Sinibaldi, ``The first interlaced continuum robot, devised to intrinsically follow the leader,'' \emph{PloS one}, vol.~11, no.~2, p. e0150278, 2016.

\bibitem{liu2021simultaneous}
W.~Liu, Raynald, Y.~Tian, J.~Gong, Z.~Ma, L.~Ma’ruf, and C.~Li, ``Simultaneous single-trajectory endoscopic biopsy and third ventriculostomy in pediatric pineal region tumors,'' \emph{Acta Neurologica Belgica}, vol. 121, pp. 1535--1542, 2021.

\bibitem{dupourque2019transbronchial}
L.~Dupourqu{\'e}, F.~Masaki, Y.~L. Colson, T.~Kato, and N.~Hata, ``Transbronchial biopsy catheter enhanced by a multisection continuum robot with follow-the-leader motion,'' \emph{International journal of computer assisted radiology and surgery}, vol.~14, no.~11, pp. 2021--2029, 2019.

\bibitem{calme2022towards}
B.~Calm{\'e}, L.~Rubbert, and Y.~Haddab, ``Towards a discrete snake-like robot based on sma-actuated tristable modules for follow the leader control strategy,'' \emph{IEEE Robotics and Automation Letters}, vol.~8, no.~1, pp. 384--391, 2022.

\bibitem{mandolino2022design}
M.~A. Mandolino, Y.~Goergen, P.~Motzki, and G.~Rizzello, ``Design and characterization of a fully integrated continuum robot actuated by shape memory alloy wires,'' in \emph{2022 IEEE 17th International Conference on Advanced Motion Control (AMC)}.\hskip 1em plus 0.5em minus 0.4em\relax IEEE, 2022, pp. 6--11.

\bibitem{kim2013stiffness}
Y.-J. Kim, S.~Cheng, S.~Kim, and K.~Iagnemma, ``A stiffness-adjustable hyperredundant manipulator using a variable neutral-line mechanism for minimally invasive surgery,'' \emph{IEEE transactions on robotics}, vol.~30, no.~2, pp. 382--395, 2013.

\bibitem{frasson2010sting}
L.~Frasson, S.~Ko, A.~Turner, T.~Parittotokkaporn, J.~F. Vincent, and F.~Rodriguez~y Baena, ``Sting: a soft-tissue intervention and neurosurgical guide to access deep brain lesions through curved trajectories,'' \emph{Proceedings of the Institution of Mechanical Engineers, Part H: Journal of Engineering in Medicine}, vol. 224, no.~6, pp. 775--788, 2010.

\bibitem{chikhaoui2018developments}
M.~T. Chikhaoui, A.~Benouhiba, P.~Rougeot, K.~Rabenorosoa, M.~Ouisse, and N.~Andreff, ``Developments and control of biocompatible conducting polymer for intracorporeal continuum robots,'' \emph{Annals of Biomedical Engineering}, vol.~46, pp. 1511--1521, 2018.

\bibitem{amanov2017toward}
E.~Amanov, J.~Granna, and J.~Burgner-Kahrs, ``Toward improving path following motion: hybrid continuum robot design,'' in \emph{2017 IEEE international conference on robotics and automation (ICRA)}.\hskip 1em plus 0.5em minus 0.4em\relax IEEE, 2017, pp. 4666--4672.

\bibitem{chen2010multi}
Y.~Chen, J.~H. Chang, A.~S. Greenlee, K.~C. Cheung, A.~H. Slocum, and R.~Gupta, ``Multi-turn, tension-stiffening catheter navigation system,'' in \emph{2010 IEEE International Conference on Robotics and Automation}.\hskip 1em plus 0.5em minus 0.4em\relax IEEE, 2010, pp. 5570--5575.

\bibitem{freschi2013technical}
C.~Freschi, V.~Ferrari, F.~Melfi, M.~Ferrari, F.~Mosca, and A.~Cuschieri, ``Technical review of the da vinci surgical telemanipulator,'' \emph{The International Journal of Medical Robotics and Computer Assisted Surgery}, vol.~9, no.~4, pp. 396--406, 2013.

\bibitem{thompson2009evaluation}
C.~C. Thompson, M.~Ryou, N.~J. Soper, E.~S. Hungess, R.~I. Rothstein, and L.~L. Swanstrom, ``Evaluation of a manually driven, multitasking platform for complex endoluminal and natural orifice transluminal endoscopic surgery applications (with video),'' \emph{Gastrointestinal endoscopy}, vol.~70, no.~1, pp. 121--125, 2009.

\bibitem{neumann2016considerations}
M.~Neumann and J.~Burgner-Kahrs, ``Considerations for follow-the-leader motion of extensible tendon-driven continuum robots,'' in \emph{2016 IEEE international conference on robotics and automation (ICRA)}.\hskip 1em plus 0.5em minus 0.4em\relax IEEE, 2016, pp. 917--923.

\bibitem{zhang2022survey}
J.~Zhang, Q.~Fang, P.~Xiang, D.~Sun, Y.~Xue, R.~Jin, K.~Qiu, R.~Xiong, Y.~Wang, and H.~Lu, ``A survey on design, actuation, modeling, and control of continuum robot,'' \emph{Cyborg and Bionic Systems}, vol. 2022, 2022.

\bibitem{henselmans2020memoflex}
P.~Henselmans, C.~Culmone, D.~Jager, R.~van Starkenburg, and P.~Breedveld, ``The memoflex ii, a non-robotic approach to follow-the-leader motion of a snake-like instrument for surgery using four predetermined physical tracks,'' \emph{Medical Engineering \& Physics}, vol.~86, pp. 86--95, 2020.

\bibitem{seeliger2020robotics}
B.~Seeliger and L.~L. Swanstr{\"o}m, ``Robotics in flexible endoscopy: current status and future prospects,'' \emph{Current opinion in gastroenterology}, vol.~36, no.~5, pp. 370--378, 2020.

\bibitem{swaney2012flexure}
P.~J. Swaney, J.~Burgner, H.~B. Gilbert, and R.~J. Webster, ``A flexure-based steerable needle: high curvature with reduced tissue damage,'' \emph{IEEE Transactions on Biomedical Engineering}, vol.~60, no.~4, pp. 906--909, 2012.

\bibitem{greer2019soft}
J.~D. Greer, T.~K. Morimoto, A.~M. Okamura, and E.~W. Hawkes, ``A soft, steerable continuum robot that grows via tip extension,'' \emph{Soft robotics}, vol.~6, no.~1, pp. 95--108, 2019.

\bibitem{schwehr2022toward}
T.~J. Schwehr, A.~J. Sperry, J.~D. Rolston, M.~D. Alexander, J.~J. Abbott, and A.~Kuntz, ``Toward targeted therapy in the brain by leveraging screw-tip soft magnetically steerable needles,'' in \emph{HSMR2022 (The 14th Hamlyn Symposium on Medical Robotics)}, 2022, pp. 81--82.

\bibitem{culha2017design}
U.~Culha, J.~Hughes, A.~Rosendo, F.~Giardina, and F.~Iida, ``Design principles for soft-rigid hybrid manipulators,'' in \emph{Soft Robotics: Trends, Applications and Challenges: Proceedings of the Soft Robotics Week, April 25-30, 2016, Livorno, Italy}.\hskip 1em plus 0.5em minus 0.4em\relax Springer, 2017, pp. 87--94.

\bibitem{gilday2023predictive}
K.~Gilday, T.~George-Thuruthel, and F.~Iida, ``Predictive learning of error recovery with a sensorized passivity-based soft anthropomorphic hand,'' \emph{Advanced Intelligent Systems}, p. 2200390, 2023.

\bibitem{gilbert2015concentric}
H.~B. Gilbert, J.~Neimat, and R.~J. Webster, ``Concentric tube robots as steerable needles: Achieving follow-the-leader deployment,'' \emph{IEEE Transactions on Robotics}, vol.~31, no.~2, pp. 246--258, 2015.

\bibitem{garriga2018complete}
A.~Garriga-Casanovas and F.~Rodriguez~y Baena, ``Complete follow-the-leader kinematics using concentric tube robots,'' \emph{The International Journal of Robotics Research}, vol.~37, no.~1, pp. 197--222, 2018.

\bibitem{nwafor2023design}
C.~J. Nwafor, C.~Girerd, G.~J. Laurent, T.~K. Morimoto, and K.~Rabenorosoa, ``Design and fabrication of concentric tube robots: A survey,'' \emph{IEEE Transactions on Robotics}, 2023.

\bibitem{dupont2009design}
P.~E. Dupont, J.~Lock, B.~Itkowitz, and E.~Butler, ``Design and control of concentric-tube robots,'' \emph{IEEE Transactions on Robotics}, vol.~26, no.~2, pp. 209--225, 2009.

\bibitem{wei2023coupling}
H.~Wei, G.~Zhang, S.~Wang, P.~Zhang, J.~Su, and F.~Du, ``Coupling analysis of compound continuum robots for surgery: Another line of thought,'' \emph{Sensors}, vol.~23, no.~14, p. 6407, 2023.

\bibitem{wu2016dexterity}
L.~Wu, R.~Crawford, and J.~Roberts, ``Dexterity analysis of three 6-dof continuum robots combining concentric tube mechanisms and cable-driven mechanisms,'' \emph{IEEE Robotics and Automation Letters}, vol.~2, no.~2, pp. 514--521, 2016.

\bibitem{cianchetti2014soft}
M.~Cianchetti, T.~Ranzani, G.~Gerboni, T.~Nanayakkara, K.~Althoefer, P.~Dasgupta, and A.~Menciassi, ``Soft robotics technologies to address shortcomings in today's minimally invasive surgery: the stiff-flop approach,'' \emph{Soft robotics}, vol.~1, no.~2, pp. 122--131, 2014.

\bibitem{ferroli2013advanced}
P.~Ferroli, G.~Tringali, F.~Acerbi, M.~Schiariti, M.~Broggi, D.~Aquino, and G.~Broggi, ``Advanced 3-dimensional planning in neurosurgery,'' \emph{Neurosurgery}, vol.~72, pp. A54--A62, 2013.

\bibitem{pishvar2020foundations}
M.~Pishvar and R.~L. Harne, ``Foundations for soft, smart matter by active mechanical metamaterials,'' \emph{Advanced Science}, vol.~7, no.~18, p. 2001384, 2020.

\bibitem{momeni2017review}
F.~Momeni, X.~Liu, J.~Ni \emph{et~al.}, ``A review of 4d printing,'' \emph{Materials \& design}, vol. 122, pp. 42--79, 2017.

\bibitem{do2020dynamically}
B.~H. Do, V.~Banashek, and A.~M. Okamura, ``Dynamically reconfigurable discrete distributed stiffness for inflated beam robots,'' in \emph{2020 IEEE International Conference on Robotics and Automation (ICRA)}.\hskip 1em plus 0.5em minus 0.4em\relax IEEE, 2020, pp. 9050--9056.

\bibitem{wagner2002role}
C.~R. Wagner, R.~D. Howe, and N.~Stylopoulos, ``The role of force feedback in surgery: analysis of blunt dissection,'' in \emph{Haptic Interfaces for Virtual Environment and Teleoperator Systems, International Symposium on}.\hskip 1em plus 0.5em minus 0.4em\relax IEEE Computer Society, 2002, pp. 73--73.

\bibitem{van2009value}
O.~A. Van~der Meijden and M.~P. Schijven, ``The value of haptic feedback in conventional and robot-assisted minimal invasive surgery and virtual reality training: a current review,'' \emph{Surgical endoscopy}, vol.~23, pp. 1180--1190, 2009.

\bibitem{watanabe2014force}
T.~Watanabe, T.~Iwai, Y.~Fujihira, L.~Wakako, H.~Kagawa, and T.~Yoneyama, ``Force sensor attachable to thin fiberscopes/endoscopes utilizing high elasticity fabric,'' \emph{Sensors}, vol.~14, no.~3, pp. 5207--5220, 2014.

\bibitem{grillo2018patient}
F.~W. Grillo, V.~H. Souza, R.~H. Matsuda, C.~Rondinoni, T.~Z. Pavan, O.~Baffa, H.~R. Machado, and A.~A.~O. Carneiro, ``Patient-specific neurosurgical phantom: assessment of visual quality, accuracy, and scaling effects,'' \emph{3D Printing in Medicine}, vol.~4, pp. 1--9, 2018.

\bibitem{collins1998design}
D.~L. Collins, A.~P. Zijdenbos, V.~Kollokian, J.~G. Sled, N.~J. Kabani, C.~J. Holmes, and A.~C. Evans, ``Design and construction of a realistic digital brain phantom,'' \emph{IEEE transactions on medical imaging}, vol.~17, no.~3, pp. 463--468, 1998.

\bibitem{altermatt2019design}
A.~Altermatt, F.~Santini, X.~Deligianni, S.~Magon, T.~Sprenger, L.~Kappos, P.~Cattin, J.~Wuerfel, and L.~Gaetano, ``Design and construction of an innovative brain phantom prototype for mri,'' \emph{Magnetic resonance in medicine}, vol.~81, no.~2, pp. 1165--1171, 2019.

\bibitem{breimer2015design}
G.~E. Breimer, V.~Bodani, T.~Looi, and J.~M. Drake, ``Design and evaluation of a new synthetic brain simulator for endoscopic third ventriculostomy,'' \emph{Journal of Neurosurgery: Pediatrics}, vol.~15, no.~1, pp. 82--88, 2015.

\end{thebibliography}

\end{document}